# *YOLOv8 to YOLO11: A Comprehensive Architecture In-depth Comparative Review*

A PREPRINT

**Priyanto Hidayatullah[a*], Nurjannah Syakrani[b], Muhammad Rizqi Sholahuddin[c], Trisna Gelar[d], Refdinal Tubagus[e]**

[abcd] Computer Engineering and Informatics Department, Politeknik Negeri Bandung, Kab. Bandung Barat, Indonesia
[e] Stunning Vision AI, Kota Cimahi, Indonesia

priyanto@polban.ac.id[a], nurjannahsy@jtk.polban.ac.id[b], muhammad.rizqi@polban.ac.id[c], trisna.gelar@polban.ac.id[d], refdinal@stunningvisionai.com[e]

**Corresponding Author*

May 4, 2026

## ABSTRACT

In the field of deep learning-based computer vision, YOLO is revolutionary. With respect to deep learning models, YOLO is also the one that is evolving the most rapidly. Unfortunately, not every YOLO model possesses scholarly publications. Moreover, there exists a YOLO model that lacks a publicly accessible official architectural diagram. Naturally, this engenders challenges, such as complicating the understanding of how the model operates in practice. Furthermore, the review articles that are presently available do not investigate the specifics of each model. The objective of this study is to present a comprehensive and in-depth architecture comparison of the four most recent YOLO models, specifically YOLOv8 through YOLO11, thereby enabling readers to quickly grasp not only how each model functions, but also the distinctions between them. To analyze each YOLO version's architecture, we meticulously examined the relevant academic papers, documentation, and scrutinized the source code. The analysis reveals that while each version of YOLO has improvements in architecture and feature extraction, certain blocks remain unchanged. The lack of scholarly publications and official diagrams presents challenges for understanding the model's functionality and future enhancement. Future developers are encouraged to provide these resources.

## 1 Introduction

YOLO (You Only Look Once) is one of the state-of-the-art models in deep learning with the most development versions, from YOLOv1 in 2015 to 2024, it has been circulating and the 11th version of YOLO is in use. Notwithstanding the success of YOLO adoption among researchers and developers, challenges persist in comprehending its architecture and components. A significant obstacle is the absence of academic publications and comprehensive architectural diagrams for certain YOLO versions, compelling researchers and practitioners to depend on secondary sources and empirical analysis to understand these models. Moreover, the existing review papers frequently lack depth in architectural specifics and provide limited insights. These challenges underscore the necessity for a thorough comparison to close the knowledge gap. A thorough architectural comparison is essential for understanding the functionality of each model and recognizing their distinct improvements.

This study seeks to rectify the deficiencies by offering a comprehensive and systematic comparison of YOLOv8 to YOLO11. This study aims to elucidate the functioning of each model, thereby augmenting comprehension of YOLO's progression. This study advances the objective of promoting innovation and development in computer vision.

## 2 Literatur Review

The YOLO series has evolved considerably since its inception, with each version presenting important enhancements and innovations aimed at improving performance and efficiency.

- YOLOv8 (2023) introduced an anchor-free detection approach, which simplifies the model architecture and enhances performance on small objects. The unified framework for multiple computer vision tasks enhances its applicability across various domains [1]. YOLOv8 has tasks that include object detection, instance segmentation, and image classification with the contribution of anchor-free detection. It was first published by Ultralytics in 2023. The current YOLOv8 supports a wide variety of visual AI tasks and added new features which are object tracking, pose estimation, and oriented bounding boxes [2].
- YOLOv9 introduced the PGI (Programmable Gradient Information) framework and GELAN (Generalized Efficient Layer Aggregation Network) architecture in 2024. They solve the problem of information bottlenecks and improve the accuracy of lightweight models [3].
- YOLOv10 (2024) is a big step forward because it uses NMS-free (non-maximum suppression) training, which makes it easier to deploy from start to finish and reduces the amount of computing needed [4]. Spatial-channel decoupled downsampling and large-kernel convolutions have made performance and efficiency a lot better [4]. The dual assignments strategy for NMS-free training consistently enhanced accuracy, decreased latency, and supported real-time object detection [4]. This model was developed by Tsinghua University researchers.
- YOLO11, the latest YOLO model from Ultralytics, delivers superior performance in various computer vision tasks, including object detection, feature extraction, instance segmentation, pose estimation, tracking, and classification. They also represent a major breakthrough in real-time object detection technology [5]. According to [6], YOLOv11 adds the C3k2 block, SPPF (Spatial Pyramid Pooling - Fast), and C2PSA (Convolutional Block with Parallel Spatial Attention) components, which improve the ability to extract features and find objects. Improving computer vision tasks like instance segmentation, pose estimation, and oriented object detection has made it more useful in a wider range of situations [7].

Numerous review articles have examined the development and progress of YOLO architectures. [8] provides a comprehensive analysis of the evolution of YOLO, focusing on versions 1 to 8 and the latest YOLO-NAS. They provide a concise historical background, examine the speed-accuracy trade-off, and emphasize the significance of application needs in selecting a YOLO version. Nonetheless, their research is limited to YOLOv8, and they do not thoroughly explore the architectural details of each iteration. [6] presents a comprehensive assessment encompassing versions 1 to 11, whereas [7] focus on versions 1 to 10, specifically highlighting the agriculture sector. [9] also address versions 1 to 10, providing a succinct summary of the architectural progression. [10] explicitly analyzes YOLOv5, YOLOv8, and YOLOv10, whereas [5] concentrate solely on YOLOv11.

[5] give a thorough look at the YOLOv1 through YOLOv10 variations, focusing on how they can be used to make photovoltaic (PV) defect detection better. As YOLO architectures get better over time, they carefully look at how these changes affect quality control in the photovoltaic sector. There are programmable gradient information, path aggregation networks, and generalized efficient layer aggregation topologies that are talked about in the paper as the main things that made each variant work. A major trend in future research, according to the review, is to improve YOLO variations to cover a wider range of PV fault scenarios. Current discussions primarily concentrate on identifying micro-cracks, but there is a recognized potential for further expansion. Attention processes in the YOLO architecture will be looked into by researchers, who know that they could greatly improve detection skills, especially for subtle and complicated defects.

Nonetheless, these reviews often lack a thorough examination of the most recent YOLO versions (v8 to v11). [8] give a thorough look at how YOLO has changed from the first version to YOLOv8. However, they neglect to examine the intricacies of the subsequent architectures, which significantly enhance the design and training of networks. [7] also do not give a full architectural comparison of YOLOv10 with its predecessors or successors, which makes it harder to understand what makes it unique and what improvements it makes. [6] give a general overview of YOLOv11, but they don't go into detail about its architectural parts and how they affect performance.

Because this study compares and contrasts YOLOv8 and YOLOv11 in a focused and thorough way, it fixes these problems by focusing on the main differences and architectural changes between the two models. By looking at the architectural details of each version and analyzing the source code, this study aims to make recent improvements easier to understand and use in real life.

# 3 Research Methods

To accurately draw the architecture schematic, we read carefully the corresponding literature. However, only reading the literature is not sufficient to understand comprehensively how the model works. Therefore, we deep dive into the source code of every YOLO version. The .yaml file contains the overall architecture blocks whereas the Python source code contains architecture blocks implementation which are in the form of Python modules.

We clone the source code from every YOLO version GitHub repository. We analyzed the YOLO overall architecture of every YOLO version from the corresponding .yaml file which is located in the ultralytics/cfg/models/<vX> folder where vX is the YOLO version. As for YOLO architecture blocks, we use conv.py, block.py, and head.py which are located in the ultralytics/nn/modules folder, except for YOLOv9. YOLOv9 .yaml file is located in the models/detect folder whereas its architecture blocks are located in models/detect/common.py. Subsequently, we created the YOLO overall architecture and architecture blocks diagrams.

# 4 Results and Discussions

In this section, we elaborate on every YOLO version's overall architecture and its blocks based on the corresponding paper, source code, and technical documentation.

## 4.1 YOLO Overall Architecture

The YOLO architecture generally consists of 3 parts: backbone, neck, and head. An essential part of the YOLO architecture, the backbone is in charge of extracting features from the input image at multiple scales. Convolutional layers and specialized blocks are stacked in this procedure to create feature maps at various resolutions. Features at different scales are integrated into the neck and transmitted to the head for prediction. Usually, this procedure entails concatenating and upsampling feature maps from many levels, allowing the model to effectively capture multi-scale data. The head generates the final predictions for object detection and classification. Bounding boxes and class labels for the objects in the image are finally output after processing the feature maps that were provided from the neck [5].

The stem, downsampling layers, stages with fundamental building blocks, and head are the elements that constitute YOLO [4]. The first component of the network, known as the stem, is in charge of taking in and processing raw input data, like pictures or videos. To improve computational efficiency and broaden the model's receptive field, the downsampling layer has been assigned to lower the spatial resolution of the features. Each stage of YOLO's architecture typically consists of multiple convolutional blocks known as basic building blocks or structures. Usually, each step produces features at a higher level of abstraction and processes data from a different resolution.

### 4.1.1 YOLOv8

The diagram is created based on yolov8.yaml which is located at ultralytics/cfg/models/v8 (Glenn Jocher & Paula Derrenger, 2023). The YOLOv8 variant is determined by three parameters: depth_multiple, width_multiple, and max_channels. The depth_multiple parameters determine how many bottleneck blocks are in the C2f block. The width_multiple and max_channels parameters define the output channels. The stem component of YOLOv8 consists of two convolution blocks with stride 2 and kernel size 3. These two blocks convert data into initial features and reduce input resolution.

The stage component in YOLOv8 uses the C2f block. There are 8 stages, which are blocks no. 2, 4, 6, 8, 12, 15, 18, and 21. Stages in the backbone (blocks no. 2, 4, 6, and 8) use shortcuts whereas the neck (blocks 12, 15, 18, and 21) does not use shortcuts. Using or not using shortcuts is guided by empirical results from trial and error experiments to find the best [13]. YOLOv8's downsampling procedure makes use of a convolution block with stride 2 and kernel size 3. When using stride 2, the output spatial resolution will be reduced by half.

SPPF (Spatial Pyramid Pooling Fast) block at the neck, right after the final block on the backbone, serves to provide a multi-scale representation of the feature map. SPPF enables the model to capture features at various levels of abstraction by pooling at different scales [14]. There are some concat and upsample blocks on the neck. The feature map's resolution is increased through upsampling. The nearest neighbor upsampling technique is used by YOLOv8. To fill the newly created pixels in a larger feature map, this technique repeats the values of neighboring pixels. Concat is used for combining feature maps. The resolution remains unchanged, but the total number of channels will increase when feature maps are combined.

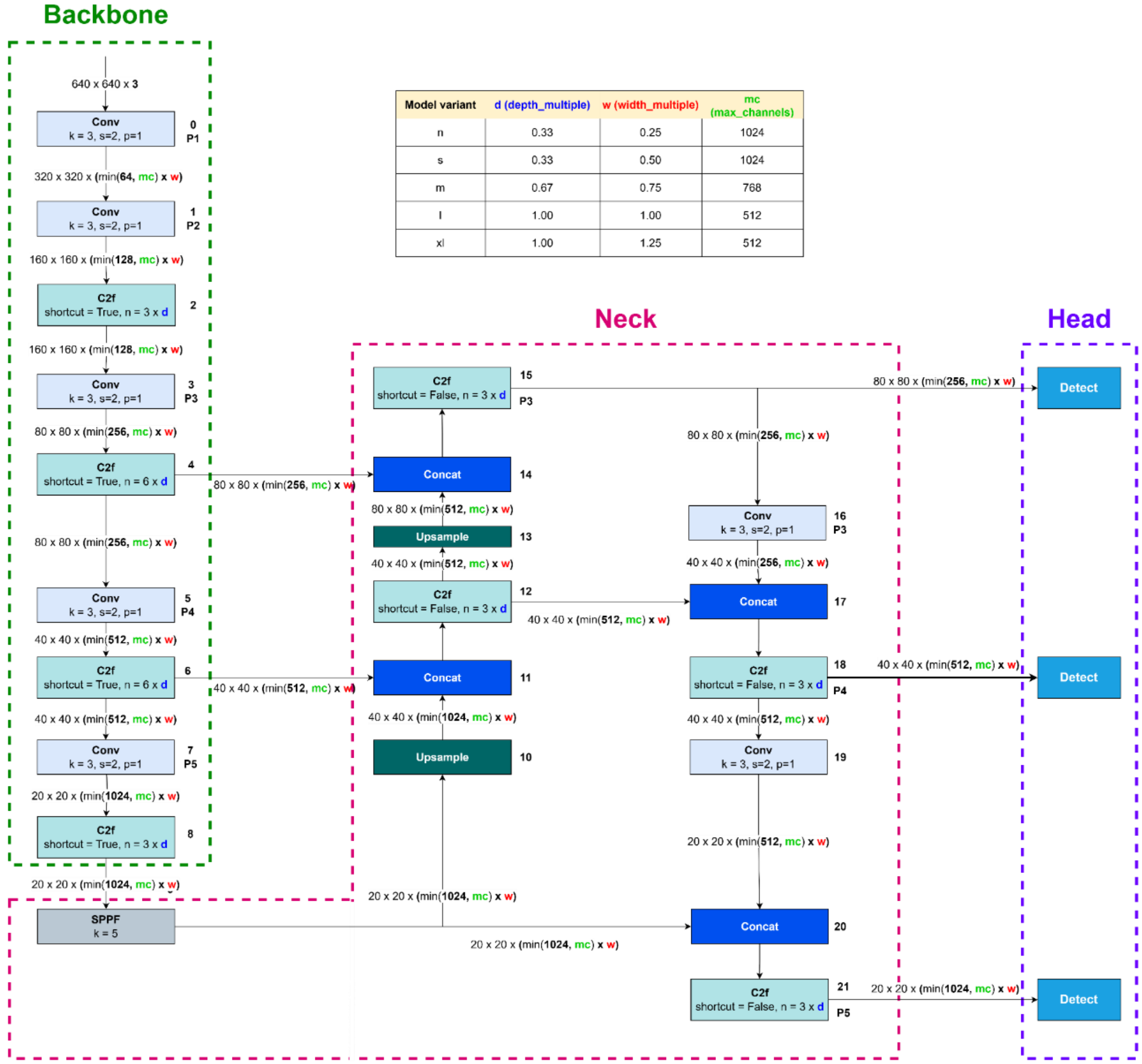

| Model variant | d (depth_multiple) | w (width_multiple) | mc (max_channels) |
|---|---|---|---|
| n | 0.33 | 0.25 | 1024 |
| s | 0.33 | 0.50 | 1024 |
| m | 0.67 | 0.75 | 768 |
| l | 1.00 | 1.00 | 512 |
| xl | 1.00 | 1.25 | 512 |

Figure 1: YOLOv8 Architecture (Adapted from [11], [12])

On YOLOv8, there are three heads. The first head, which is connected to block No. 15, is used to detect small objects. The second head, which is connected to block No. 18, is used to detect medium objects. The third head, which is connected to block No. 21, is used to detect large objects. As an important note, in every YOLO version, the object size is relative to the image or video frame.

### 4.1.2 YOLOv9

The diagram is created based on yolov9-c.yaml which is located at models/detect/ folder [15]. In YOLOv9, there is one additional section called the auxiliary. This section is used to generate reliable gradients and update network parameters. By providing additional information that connects the input data with the output target, the problem of information loss when passing through the layers of a deep learning network can be resolved [3]. This auxiliary section is only used in the training process. This section can be eliminated during the inference process to speed up the model without sacrificing accuracy.

YOLOv9 starts with a Silence block. The input is simply returned by this block; no transformation operations are carried out. This block connects the YOLOv9 input to the auxiliary and backbone sections. The stem component in YOLOv9 is two convolution blocks with kernel size 3 and stride 2. These two blocks convert data into initial features and reduce input resolution.

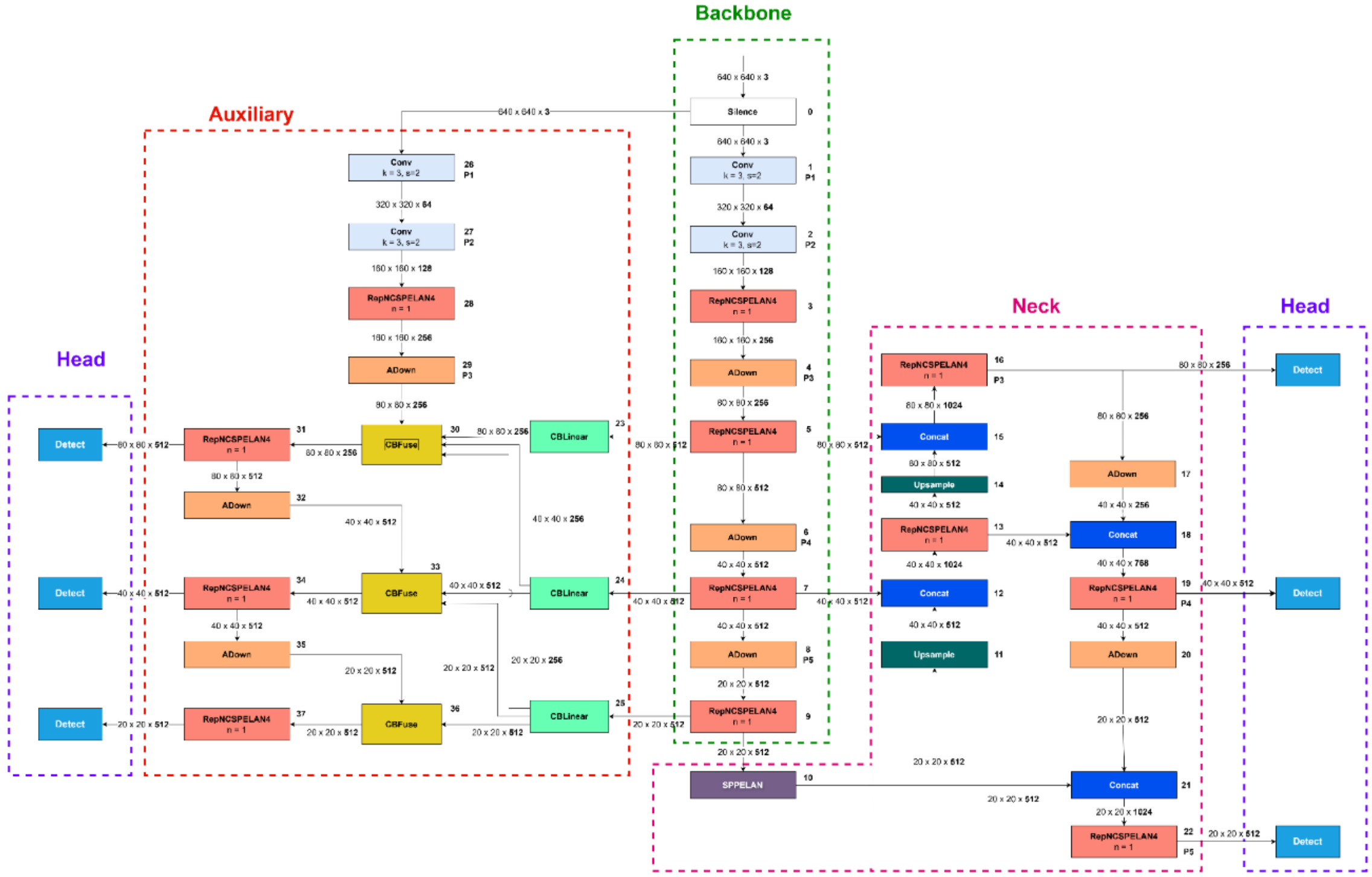

Figure 2: YOLOv9-c Architecture (Adapted from [16])

The stage component in YOLOv9 uses the RepNCSPELAN4 block. RepNCSPELAN4 is an implementation of the GELAN concept in YOLOv9. GELAN combines CSPNet and ELAN (on YOLOv7) designed with gradient path planning, taking into account lightweight, inference speed, and accuracy. GELAN generalizes ELAN capabilities. ELAN only uses stacking of convolutional layers whereas GELAN can use various computing blocks such as residual blocks, or more complex blocks (C.-Y. Wang et al., 2024). There are 8 stages in YOLOv9 which are blocks 3, 5, 7, 9, 13, 16, 19, and 22. The downsampling process in YOLOv9 uses the ADown block.

At the neck, precisely after the last block on the backbone, there is a SPPELAN block. This block is the same as SPPF (Spatial Pyramid Pooling Fast) in YOLOv8. There are also some upsample and concat blocks. Just like in YOLOv8, YOLOv9 uses nearest neighbor upsampling.

There are 3 heads in YOLOv9. The purpose of the first head, which is attached to block No. 16, is to detect small objects relative to the frame or image. Detecting medium objects is done by the second head, which is attached to block No. 19. For detecting large objects, the third head is used, which is connected to block No. 22.

The auxiliary section provides additional information that links the input data to the target output. Therefore, this section takes input from the Silence block and adds several blocks that are identical to those in the backbone which are two convolutions, RepNCSPELAN4, and ADown.

There are three CBLinear blocks. These blocks are used to create different pyramidal feature maps and are used to obtain higher-level features from the first backbone. The CBFuse block is the main part that contains the reversible property, which combines the high-level features of the first backbone with the low-level features of the second backbone.

Besides the auxiliary section, one novelty in YOLOv9 is additional heads. There are three additional heads. Heads connected to blocks No. 31, No. 34, and No. 37 are used to detect small, medium, and large objects correspondingly.

### 4.1.3 YOLOv10

The diagram is created based on yolov10l.yaml which is located in ultralytics/cfg/models/v10 folder [17]. YOLOv10 architecture is built based on YOLOv8. Therefore, it has a similar stem component and parameters to YOLOv8 parameters. The difference is that depth_multiple parameters in YOLOv10 is used to define how many bottleneck blocks are in C2f block and how many CIB blocks are in the C2fCIB block.

Figure 3 shows the YOLOv10l architecture. The differences between the other variants lie in the use of the C2fCIB block. Details of the C2fCIB block usage for each variant can be seen in the Table 1.

YOLOv10's stage component uses the C2f or C2fCIB block. There are eight stages, which are blocks 2, 4, 6, 8, 13, 16, 19, and 22. The concept of rank-guided block design, which aims to optimize the YOLO architecture by taking into account the complexity of the basic blocks at each stage of the model based on intrinsic rank analysis, is present in YOLOv10. The intrinsic rank is a numerical value that indicates the number of significant singular values at the end of basic blocks. There is typically more redundancy in the larger scale models and the deeper stages of the YOLO model. CIB (Compact Inverted Block) will be used to replace the bottleneck block in C2f at high redundancy stages, transforming it into C2fCIB (A. Wang et al., 2024). For instance, blocks 8, 13, 19, and 22 of the YOLOv10l variant, as shown in the above image, use C2fCIB. The Table 1 presents C2fCIB utilization in each YOLOv10 variant.

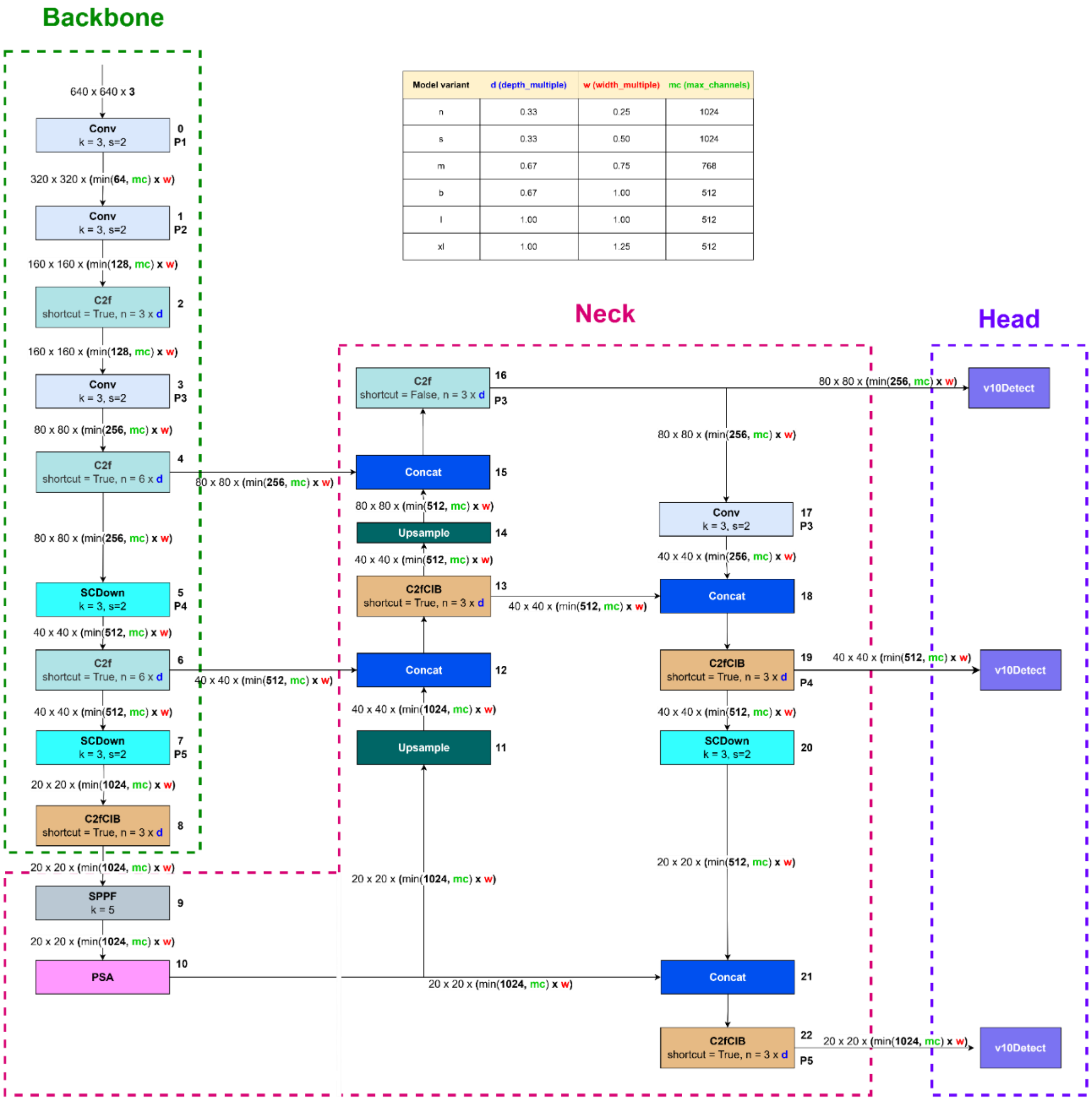

| Model variant | d (depth_multiple) | w (width_multiple) | mc (max_channels) |
|---|---|---|---|
| n | 0.33 | 0.25 | 1024 |
| s | 0.33 | 0.50 | 1024 |
| m | 0.67 | 0.75 | 768 |
| b | 0.67 | 1.00 | 512 |
| l | 1.00 | 1.00 | 512 |
| xl | 1.00 | 1.25 | 512 |

Figure 3: YOLOv10l Architecture (Adapted from [18])

Table 1: C2fCIB block usage

| Model Variant | C2fCIB |
|---|---|
| n | 22 |
| s | 8, 22 |
| m | 8, 9, 12 |
| b | 8, 13, 19, 22 |
| l | 8, 13, 19, 22 |
| x | 6, 8, 13, 19, 22 |

A Spatial Pyramid Pooling Fast (SPPF) block, analogous to the YOLOv8 architecture, is positioned within the neck immediately after the final module of the backbone. Following this, the Partial Self Attention (PSA) block is introduced. PSA combines global modeling capabilities with a self-attention mechanism to enhance performance [4]. To mitigate the excessive computational burden arising from the intricacies associated with quadratic self-attention, PSA is implemented solely after stage 4, utilizing the lowest resolution [4].

As in YOLOv8, there exist several concat and upsample blocks within the neck. The upsample block employs the nearest neighbor upsampling as well. There are three heads. Heads connected to blocks No. 16, No. 19, and No. 22 are used to detect small, medium, and large objects correspondingly.

#### 4.1.4 YOLO11

The diagram is created based on yolo11.yaml which is located in ultralytics/cfg/models/11 folder [19]. In YOLO11 there exist 3 variables that dictate the YOLO11 variant. These variables are identical to those in YOLOv8, specifically depth_multiple, width_multiple, and max_channels. The difference is the depth_multiple variable ascertains the quantity of bottleneck or C3k blocks present in block C3k2 and PSA blocks within C2PSA. The width_multiple and max_channels variables ascertain the output channels. The stem components of YOLO11 are congruent with those of YOLOv8.

The stage component in YOLO11 employs the C3k2 block. C3k2 represents an advancement of C2f in YOLOv8 and YOLOv10. C3k2 possesses a reduced number of parameters in comparison to C2f. Within the C3k2 block, there exists a c3k parameter which is utilized to ascertain the implementation of the c3k module. If True, then it will employ the C3k module. Conversely, if c3k is False, it will utilize Bottleneck as observed in C2f. However, in the variants m, l, and x, the value of c3k is consistently True. This finding is not available in the YOLO11 documentations [2]. Instead, we found it after analyzing the source code. Derived from this experience, examining the source code is essential to comprehend how the YOLO model works.

There are eight stages, specifically blocks numbered 2, 4, 6, 8, 13, 16, 19, and 22. The downsampling procedure in YOLO11 employs a convolutional block with a kernel dimension of 3 and a stride of 2. When utilizing a stride of 2, the resultant spatial resolution will be diminished by 50 percent.

On the neck part, precisely after the terminal segment of the backbone, there exists an SPPF (Spatial Pyramid Pooling Fast) block analogous to that in YOLOv8. Following this, there is the C2PSA block. Analogous to the PSA in YOLOv10, C2PSA employs a self-attention mechanism, which is utilized to enhance efficacy by combining global modeling capabilities. C2PSA is exclusively positioned after stage 4 with minimal resolution. As in YOLOv8 and YOLOv10, there exist several concat and upsample blocks within the neck. The upsample block employs nearest-neighbor upsampling as well. The head part of YOLO11 is congruent with those of YOLO10, in terms of their functionality and block number.

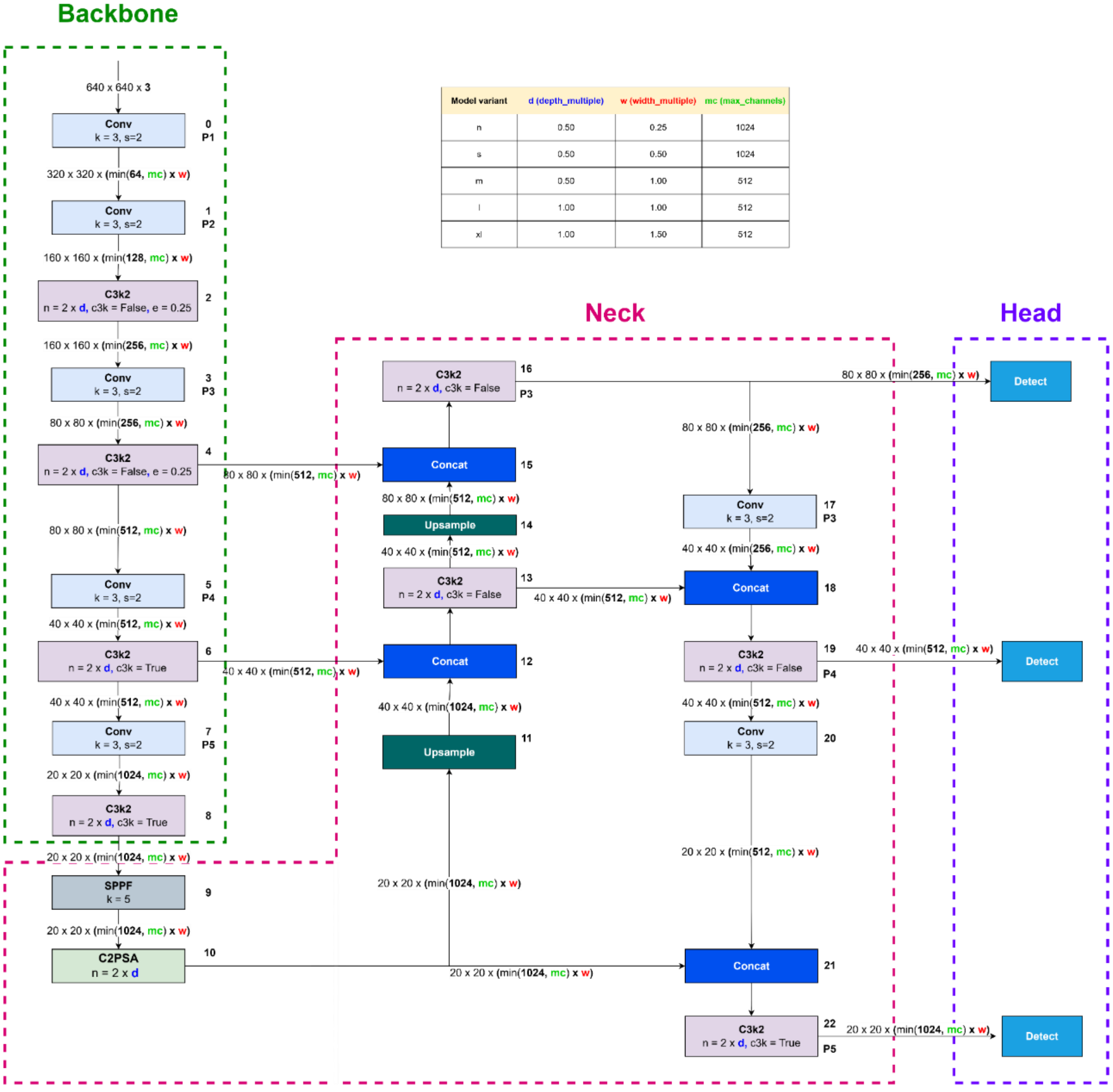

| Model variant | d (depth_multiple) | w (width_multiple) | mc (max_channels) |
|---|---|---|---|
| n | 0.50 | 0.25 | 1024 |
| s | 0.50 | 0.50 | 1024 |
| m | 0.50 | 1.00 | 512 |
| l | 1.00 | 1.00 | 512 |
| xl | 1.00 | 1.50 | 512 |

Figure 4: YOLO11 Architecture (Adapted from [20])

### 4.2 YOLO Architecture Blocks

The architecture blocks drawing of YOLOv8 is adapted from [11], [12] whereas YOLOv9 to YOLO11 architecture blocks drawings are adapted from [16], [18], [20].

#### 4.2.1 Input Image Resizing

In contrast to YOLOv4 and its predecessors, the input image in YOLOv8 to YOLO11 will be resized while preserving the image aspect ratio. To preserve the aspect ratio, the image will be padded with gray pixels. If the input image is a square, the image will be resized without padding. Figure 5 illustrates the example of the image resizing process.

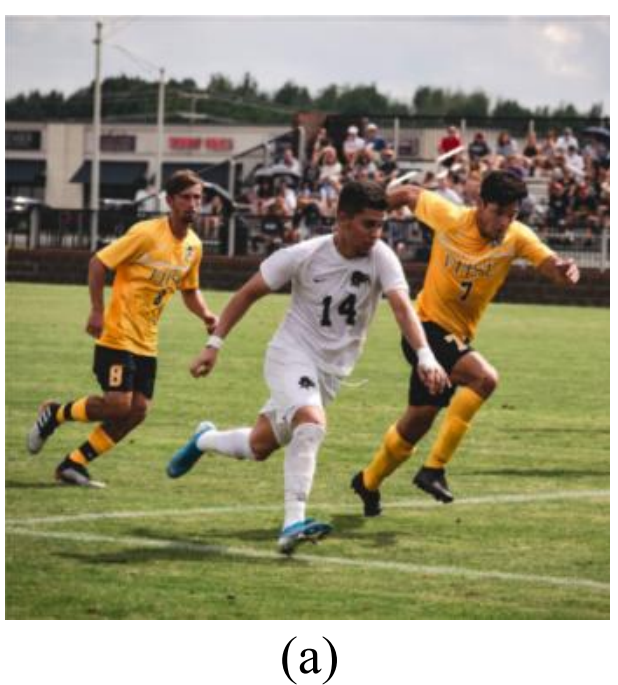
(a)

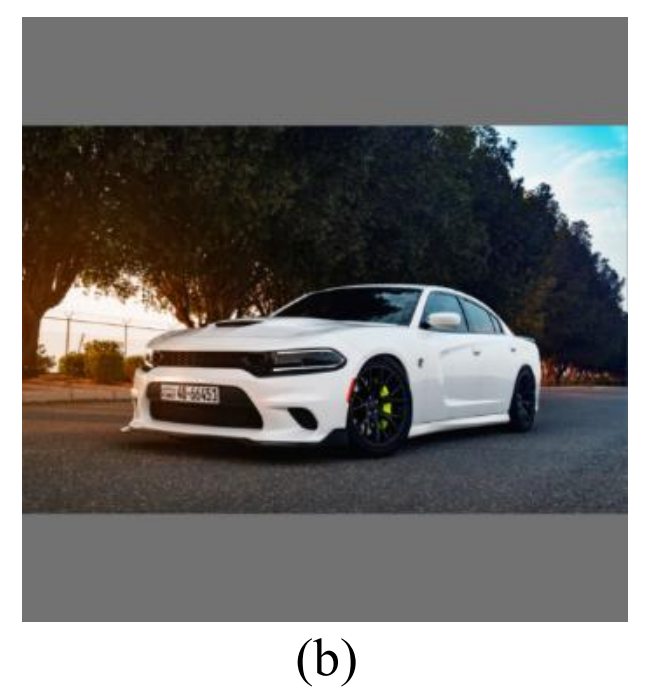
(b)

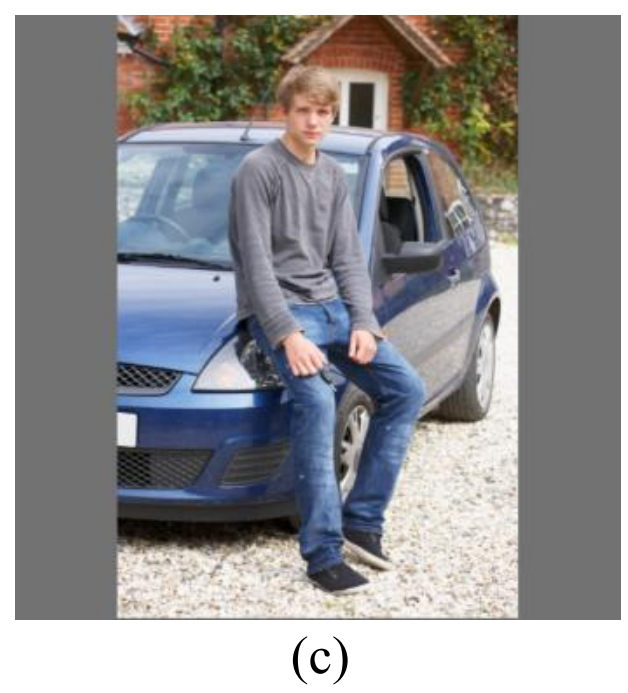
(c)

Figure 5: Image Resizing (a) Square Image (b) Landscape Image (c) Portrait Image

#### 4.2.2 Convolutional (Conv) Block

In all YOLO versions in this study, the convolutional block can be drawn in Figure 6. It contains a 2-dimensional convolutional layer, a 2-dimensional batch normalization, and a SILU activation function. They are fused into a single convolutional block. However, YOLO11 has two new convolutional blocks. The first one is a convolutional block without activation function and the second is the depthwise convolutional block (DWConv).

The feed-forward network (FFN) in PSABlock (Position-Sensitive Attention Block) uses Conv block without activation function. In depthwise convolution, only one kernel is used to operate on a single channel, which makes DWConv lighter. The matrix value of every kernel varies.

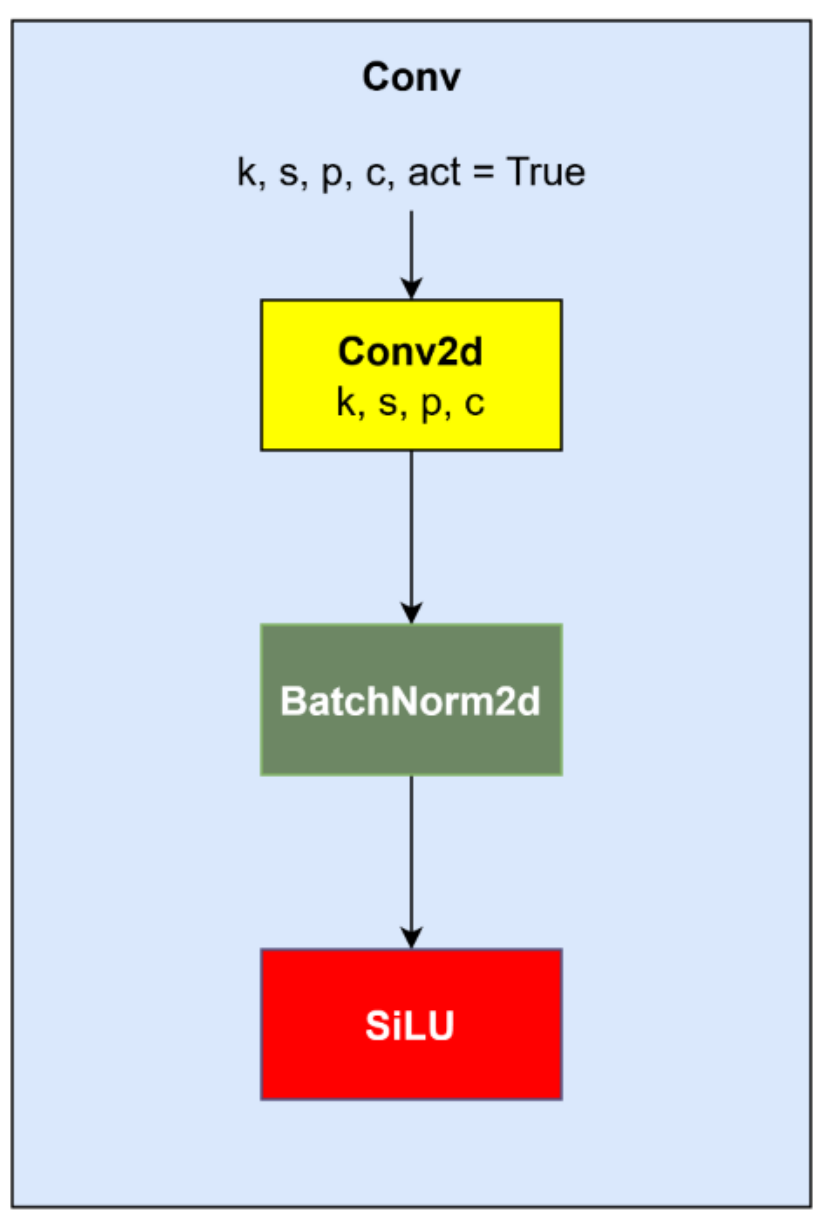

Figure 6: Standard recent YOLO convolutional block

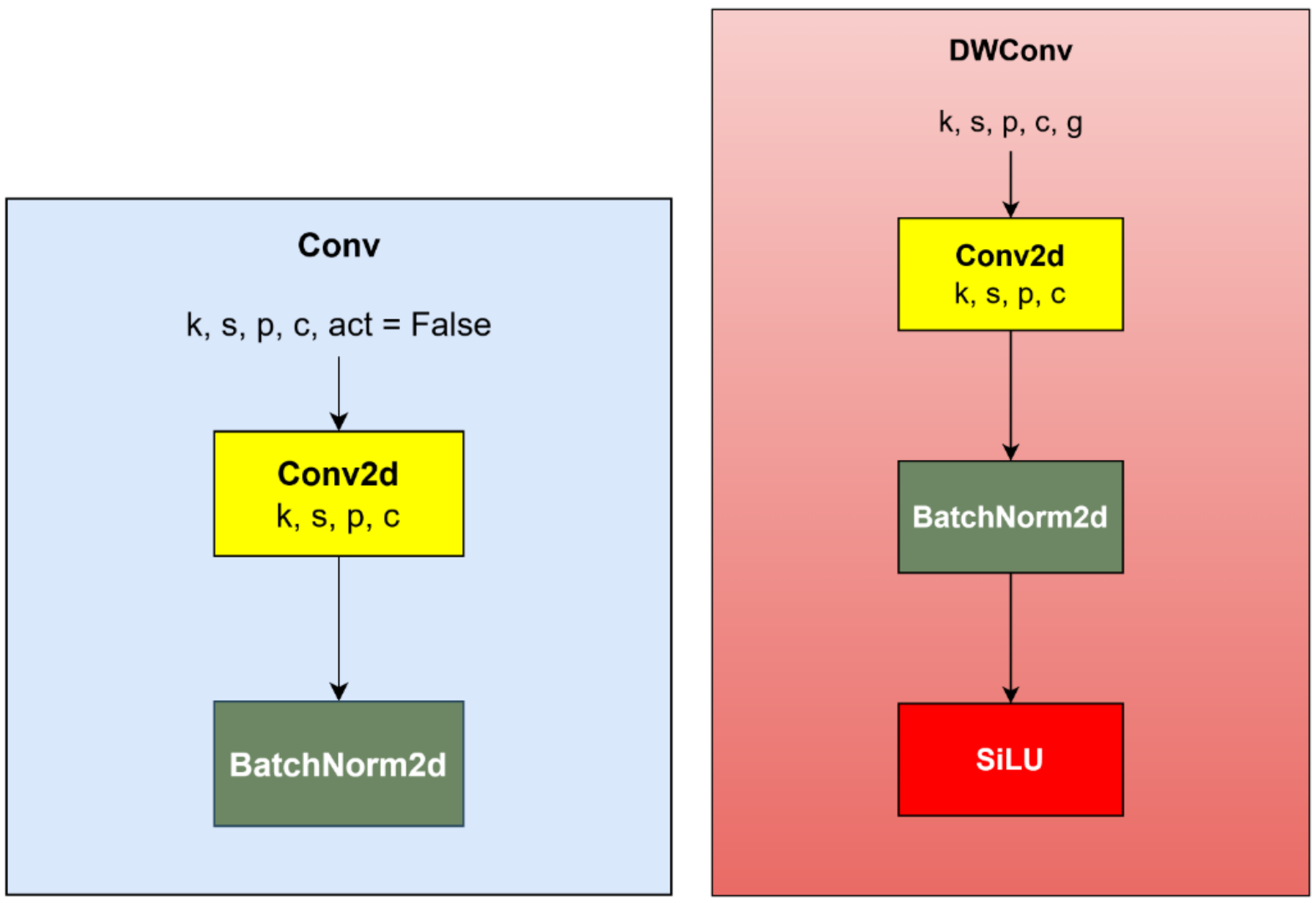

Figure 7: Two new YOLO11 convolutional blocks

#### 4.2.3 Downsampling Block

Both the YOLOv8 and YOLO11 versions employ typical 3×3 convolution with stride 2 for the downsampling process. However, this traditional method uses a lot of resources and may be less efficient due to its computing costs and relatively large number of parameters.

Spatial-Channel Decoupled Downsampling (SCDown), which separates spatial reduction and channel addition operations, is used in YOLOv10. First, the number of channels is altered without affecting the spatial resolution using pointwise convolution (1×1). The spatial resolution is then decreased without altering the number of channels via depthwise convolution (e.g. 3×3). This strategy reduces the computational cost and the number of parameters [4]. Figure 8 is the illustration of SCDown.

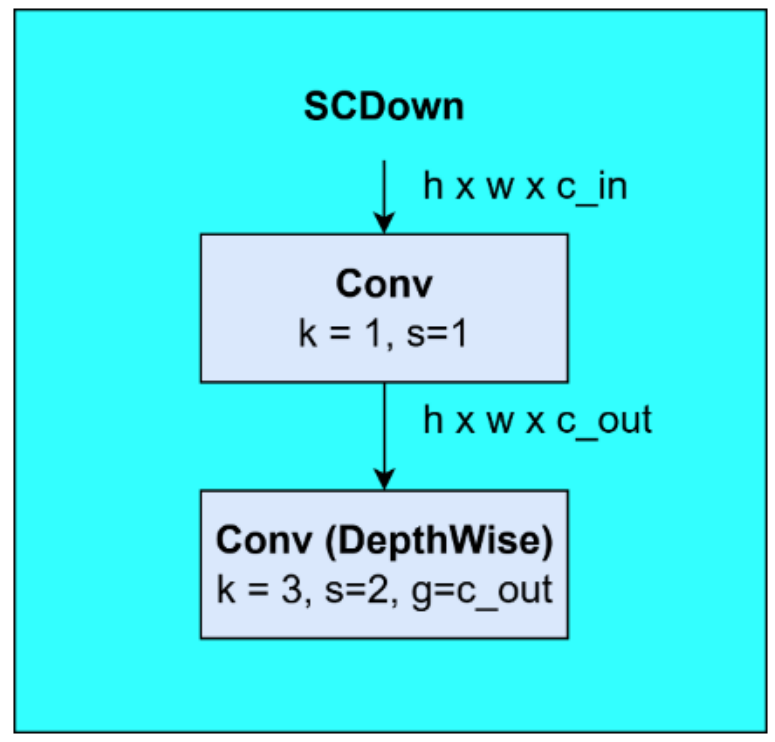

Figure 8: SCDown block in YOLOv10

YOLOv9 employs Adaptive Downsampling (ADown). This block uses a combination of Average Pooling and Maxpooling to perform downsampling. In ADown, the parameter count is comparatively lower, as pooling involves no parameters. [21]. Figure 9 is the illustration of ADown.

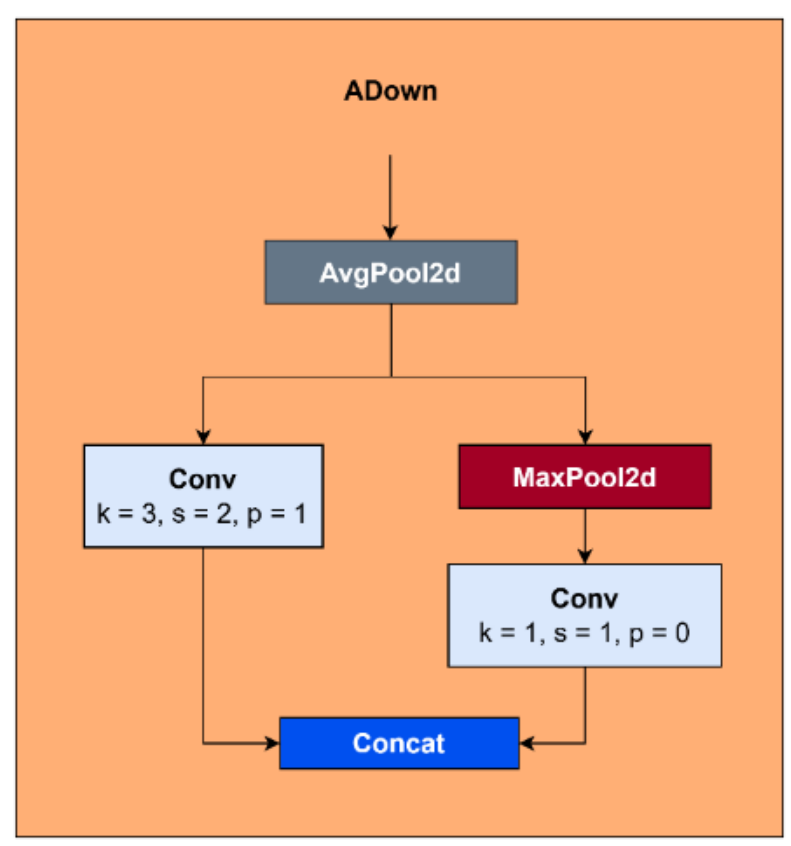

Figure 9: ADown block in YOLOv9

#### 4.2.4 Bottleneck

The bottleneck is a block that is similar to the Resnet block. This block is available in YOLOv8, YOLOv10, and YOLO11. Bottlenecks enable deeper networks by stacking additional layers with little increase in computing costs. Some bottlenecks utilize shortcuts and those that do not, which are illustrated in Figure 10.

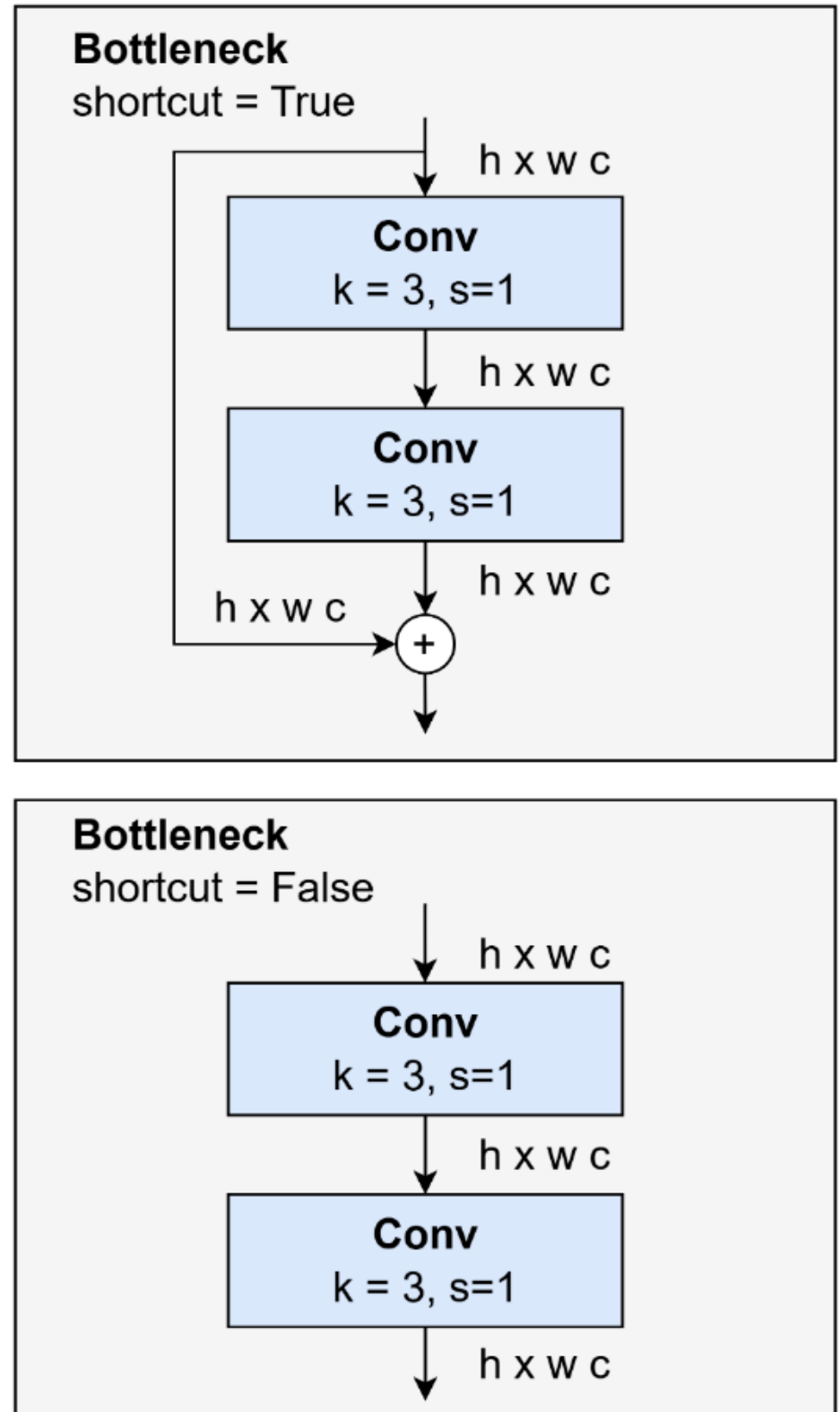

Figure 10: Bottleneck blocks

#### 4.2.5 C2f Block

C2f is a faster Implementation of CSP Bottleneck with 2 convolutions. C2f is utilized for feature extraction at all stages. C2f's modules can successfully learn multiscale features and broaden the range of receptive fields by utilizing feature vector switching and multilayer convolution. C2f is available in YOLOv8 and YOLOv10. It is illustrated in Figure 11.

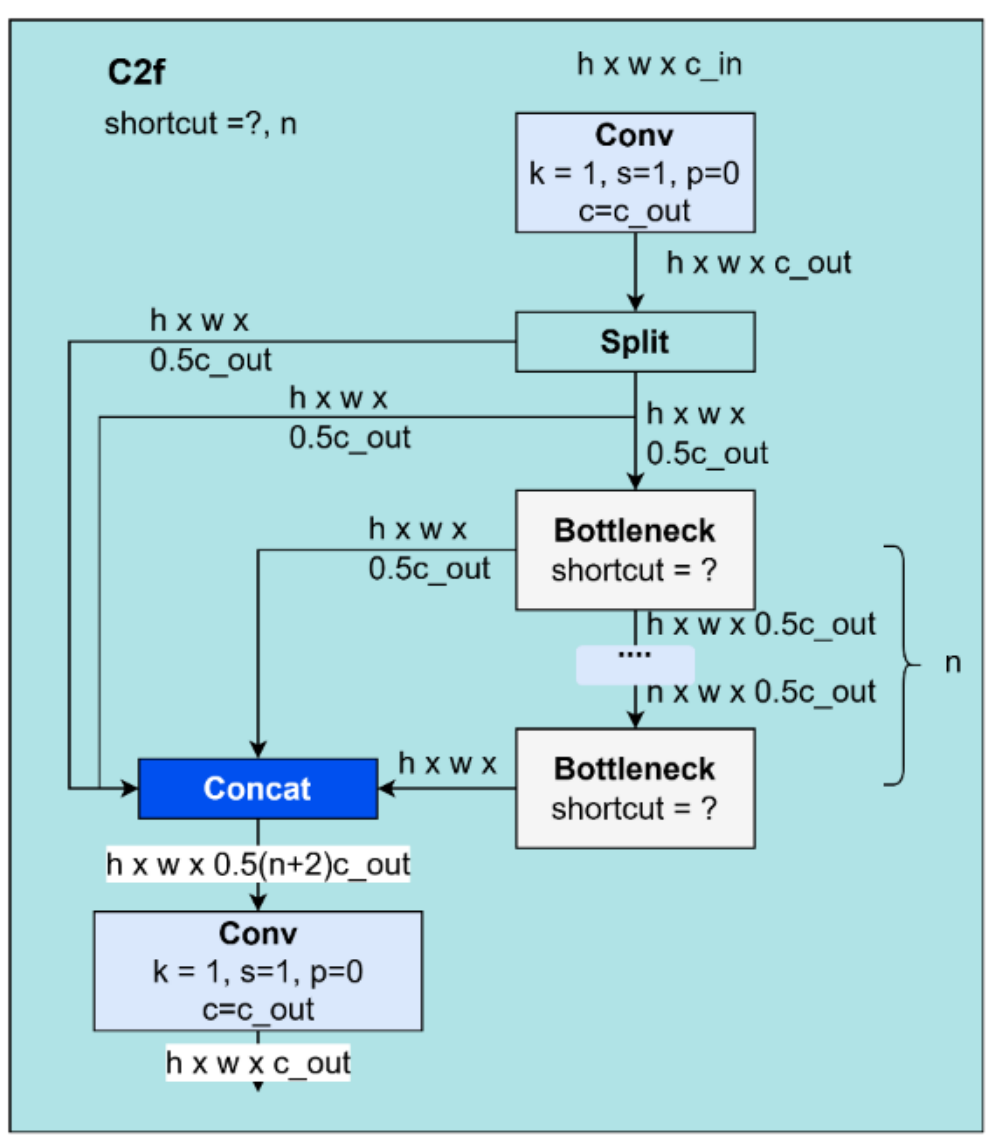

Figure 11: C2f Block

#### 4.2.6 C2fCIB, CIB, RepVGGDW

Besides C2f, YOLOv10 introduced C2fCIB. YOLOv10 employs a rank-guided block design paradigm. This is a strategy for optimizing the YOLO model architecture by modifying the complexity of the fundamental blocks at each model stage using intrinsic rank analysis. Intrinsic rank is a numerical measure that counts the number of significant singular values in the final convolution of a basic block.

The YOLO model and other large-scale models tend to have greater redundancy as they progress deeper. At stages with significant redundancy, the bottleneck block in C2f will be replaced with CIB (Compact Inverted Block) to form C2fCIB. For example, in the YOLOv10l version, C2fCIB is used in blocks 8, 13, 19, and 22. This strategy improves the model efficiency, i.e., it increases the model training speed compared to YOLOv8 [4]. Figure 12 illustrates C2fCIB block.

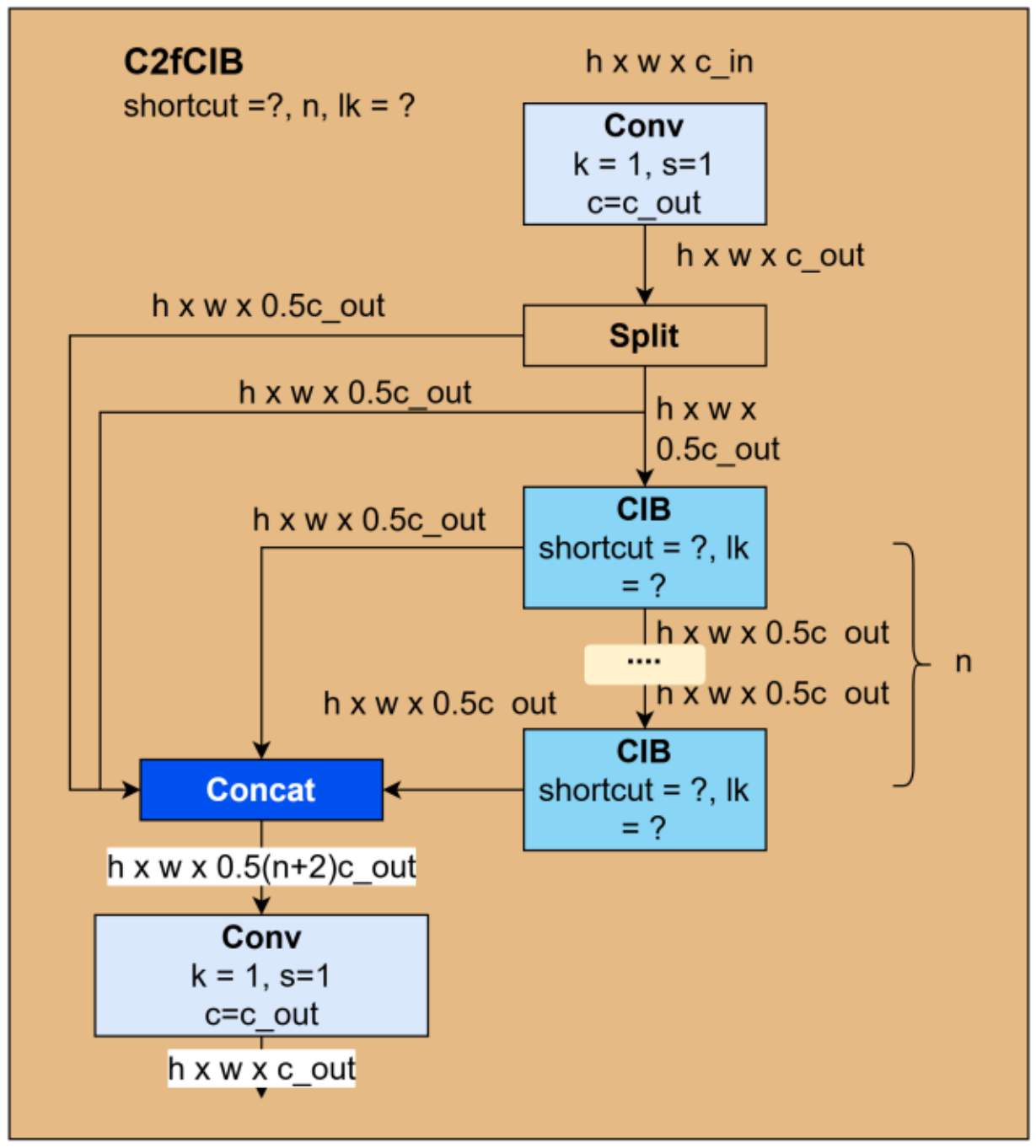

Figure 12: C2fCIB block

CIB uses depthwise convolution. The block structure is depthwise convolution, convolution, depthwise convolution, convolution, and depthwise convolution. Some CIBs use shortcuts and some do not.

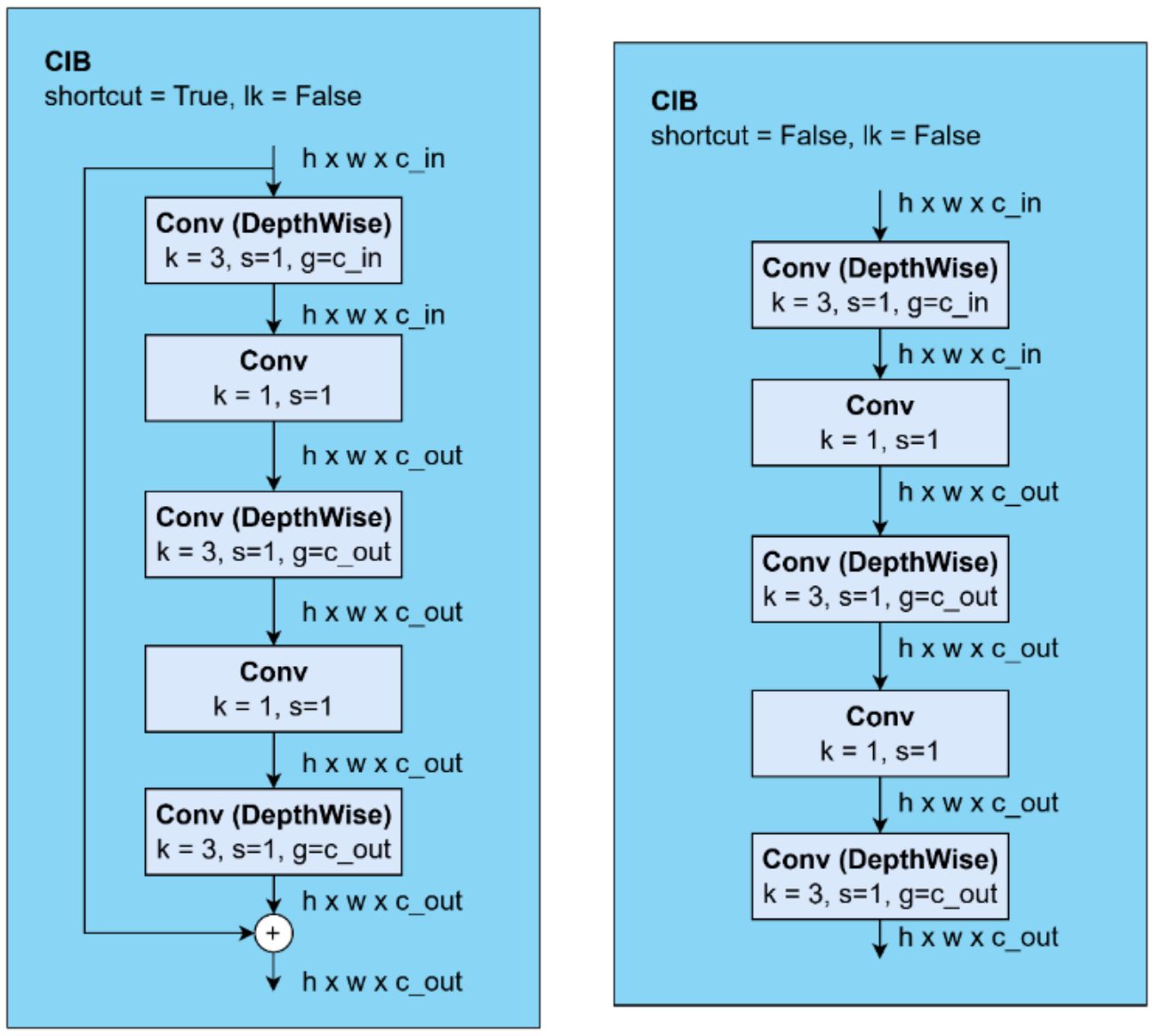

Figure 13: CIB block

The concept of large-kernel convolution is presented in YOLOv10. By using this method, the input region that affects the convolution layer's neurons' output can be expanded. Large-kernel convolution is used only in deep stages of the model. In its implementation, the second depthwise convolution in CIB is changed to RepVGGDW. RepVGGDW contains two depthwise convolutional blocks with kernel sizes of three and seven. Because there is no gain when applied to larger versions, large-kernel convolution is only utilized on the Yolov10 n and s variants.

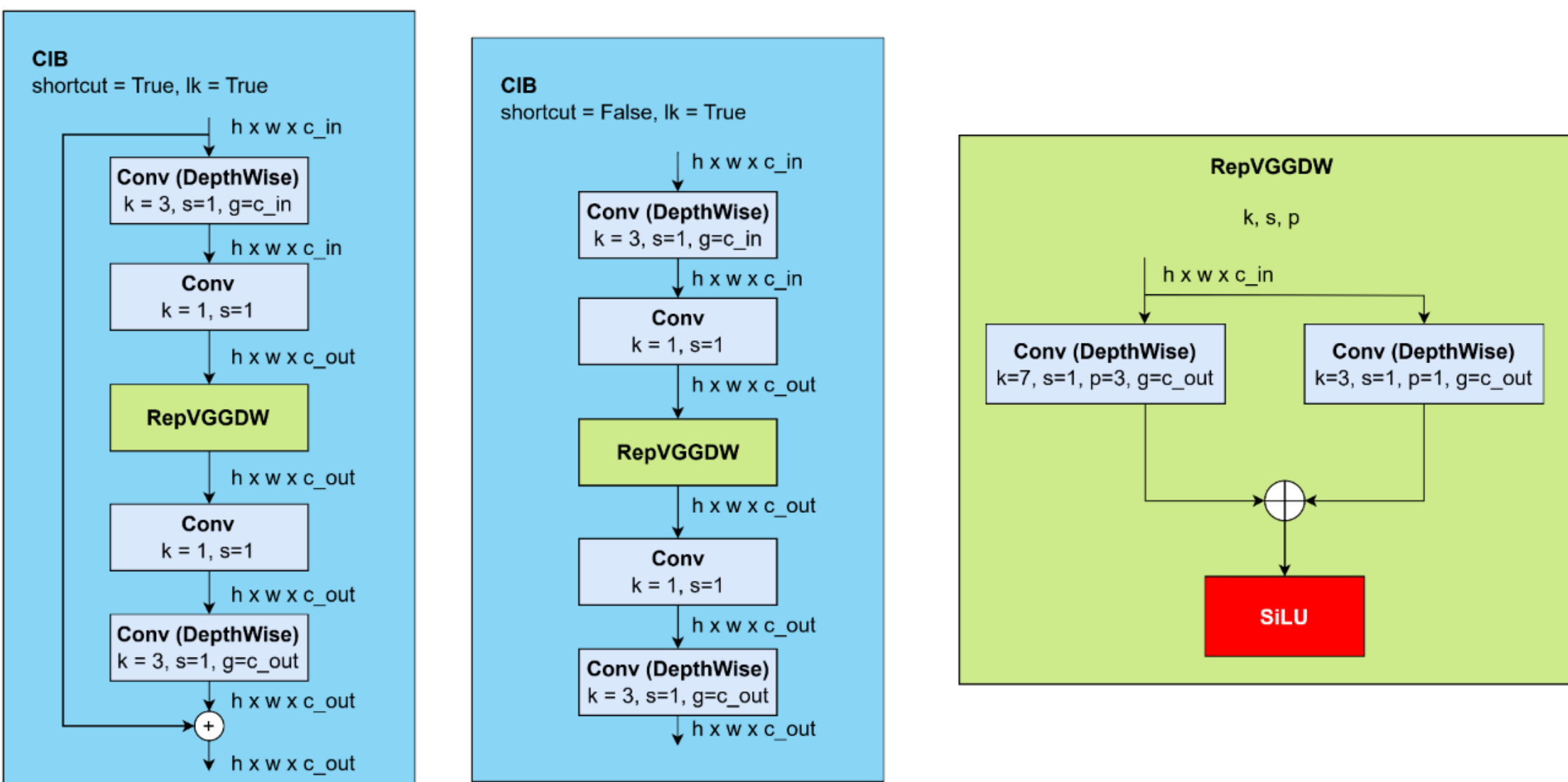

Figure 14: CIB block with RepVGGDW

### 4.2.7 C3k2 Block

This block is newly introduced in YOLO11. C3k2 is the replacement for the C2f block in YOLOv8. C3k2 has a similar structure to C2f. Moreover, the first version C3k2 is congruent with C2f. The second version of C3k2 replaced the bottlenecks with one C3k block as seen in Figure 15.

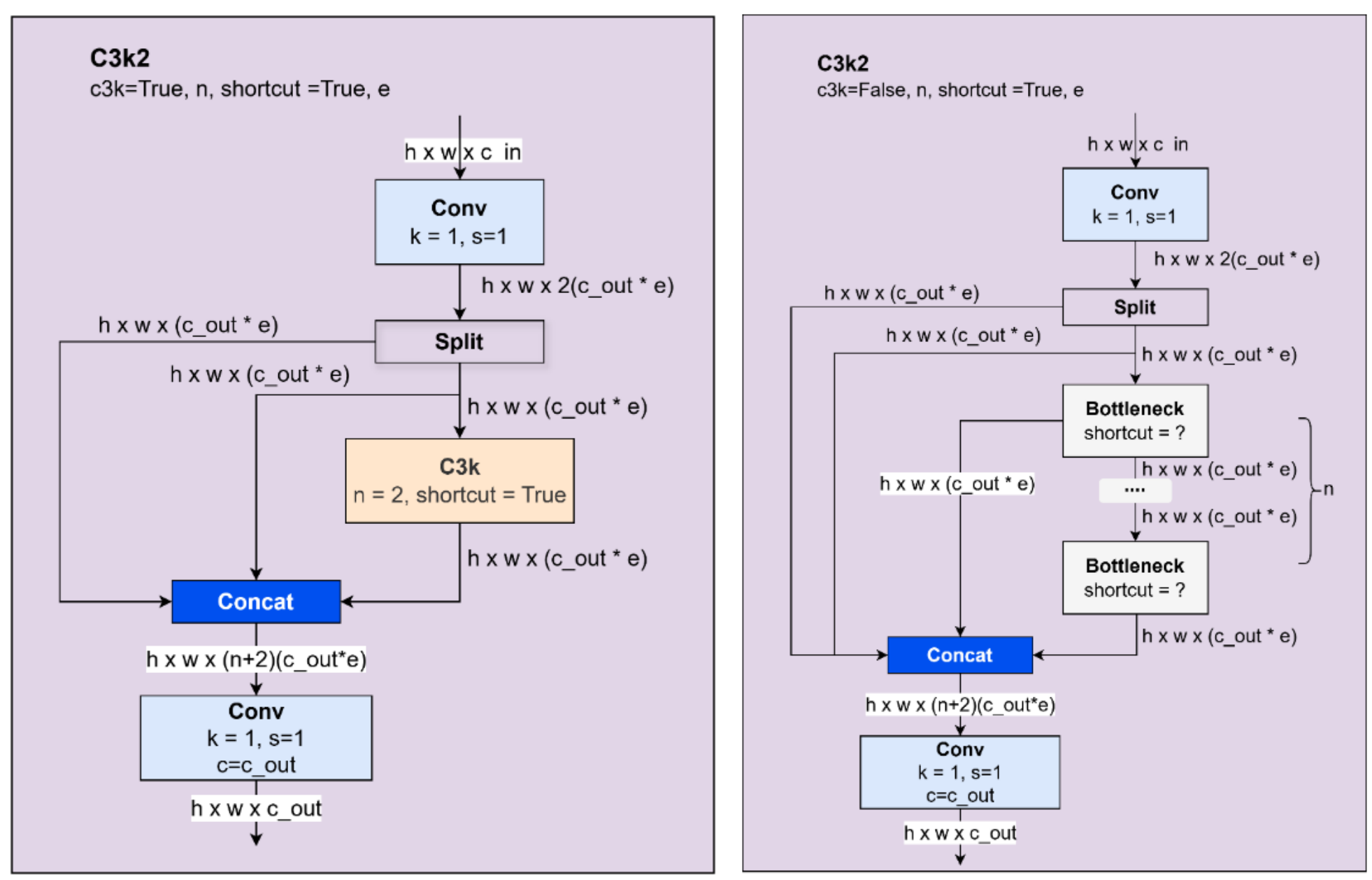

Figure 15: C3k2 blocks the first and second version

C3k itself is a block containing 3 convolution blocks and several bottlenecks. The bottleneck has two versions, one with a shortcut (shortcut=true) and one without a shortcut (shortcut=false). The C3k2 block effectively captures more complex features, making it appropriate for objects of varying sizes. The illustration for C3k is in Figure 16.

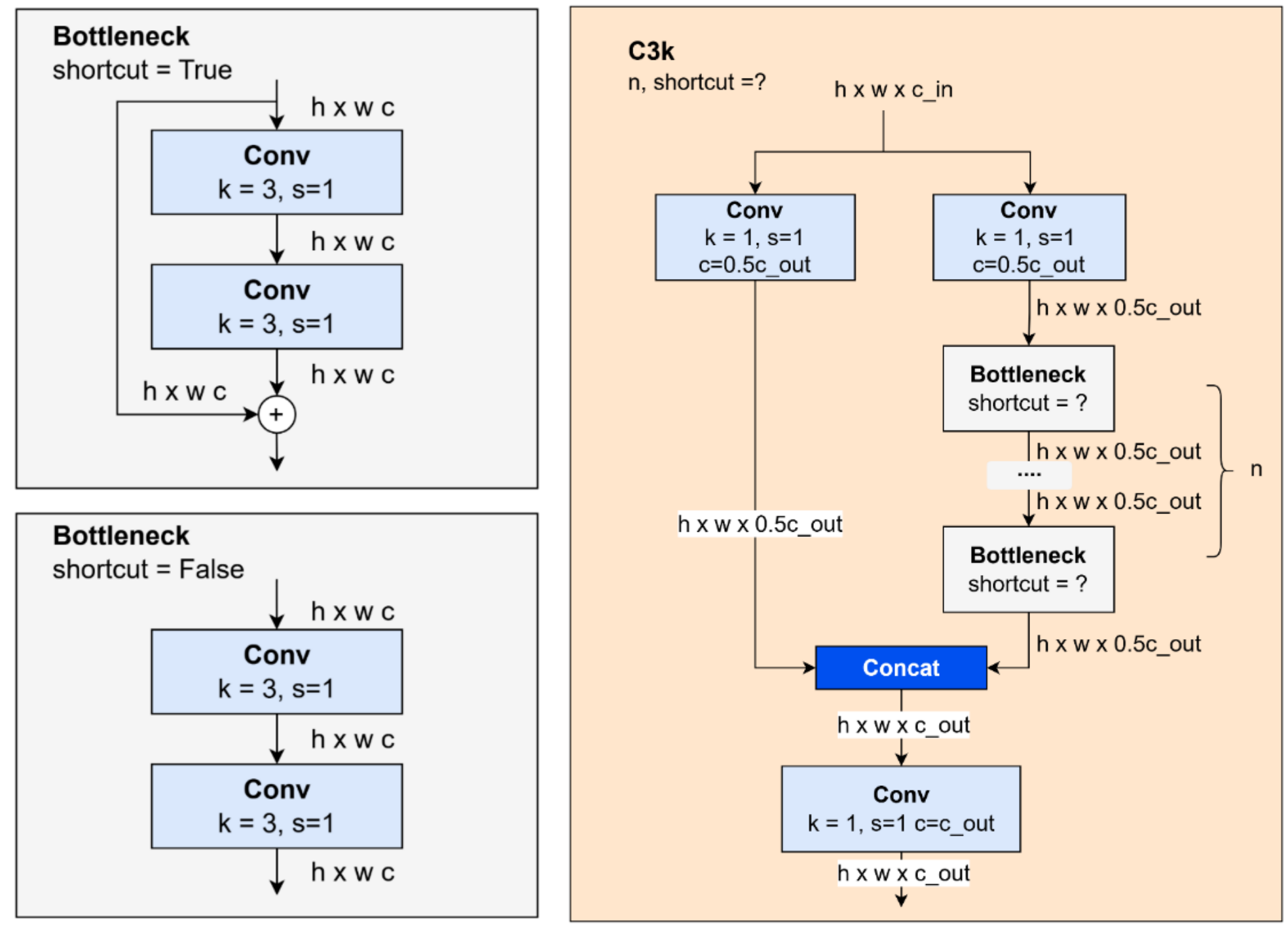

Figure 16: Bottleneck and C3k block

#### 4.2.8 SPPF & SPPELAN

Spatial Pyramid Pooling – Fast, or SPPF, constitutes a block that permits feature maps to be represented across multiple scales. By pooling at varying scales, SPPF allows the model to capture features at different levels of abstraction. This block was first introduced in YOLOv8 and later adapted for YOLOv10 and YOLO11.

SPPF represents a modification of SPP or Spatial Pyramid Pooling which is available in the YOLO model before YOLOv8. The Maxpool kernel size in SPP is 3, 5, and 9. In contrast, SPPF employs a kernel size of 5. This modification reduces the quantity of floating-point operations (FLOP) in SPPF relative to SPP. Some of the YOLOv9 blocks exhibit analogous structure and function to SPPF. However, in YOLOv9, it is named SPPELAN. Figure 17 illustrates SPPF and SPPELAN block.

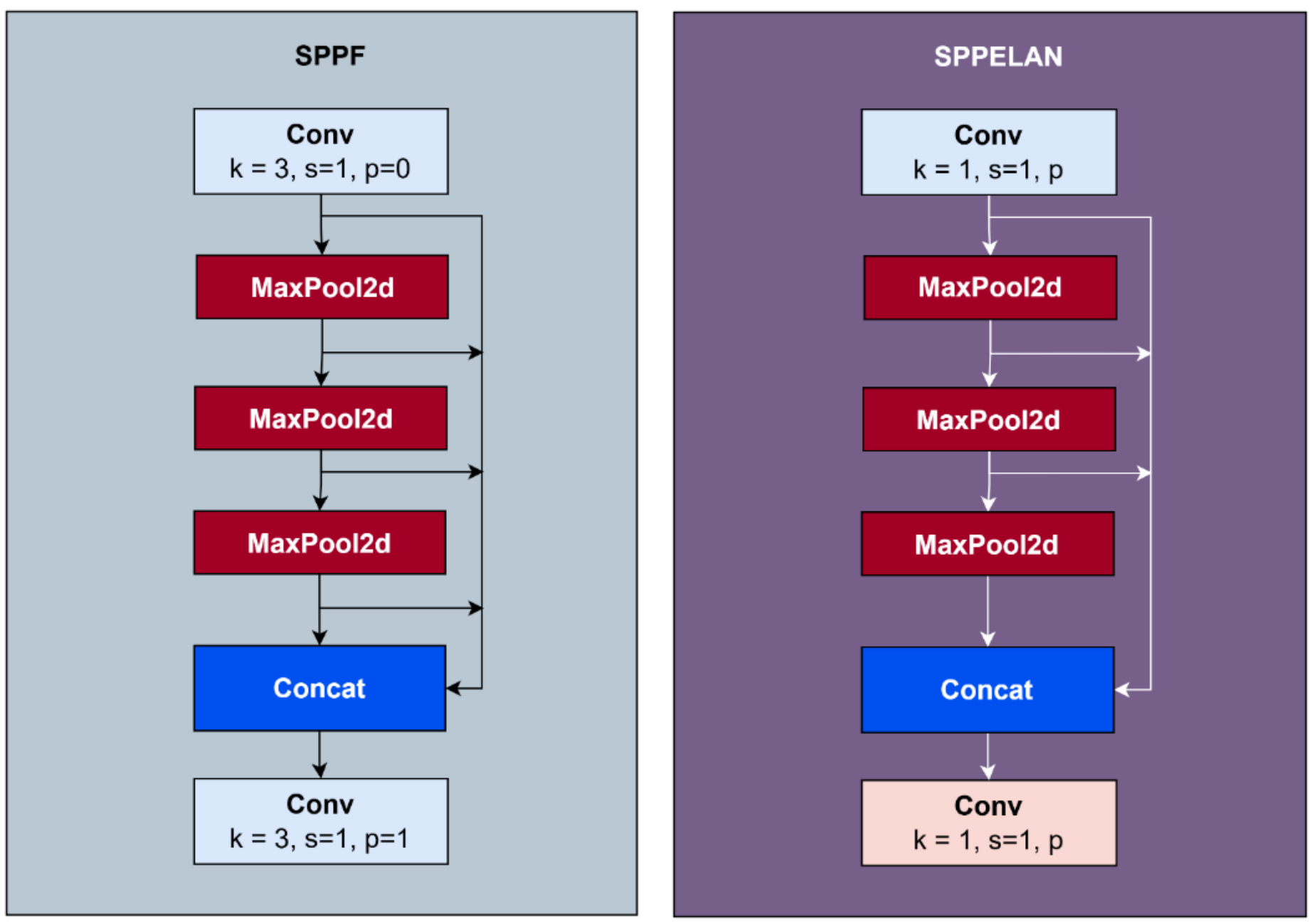

Figure 17: SPPF and SPPELAN Block

#### 4.2.9 C2PSA & PSABlock

The attention module is employed to ascertain how the pixels within an image or video frame interrelate. The query, key, and value constitute the three parameters that each pixel within the image retains. The dot product applied to all keys from other pixels will be utilized to execute queries on a particular pixel. The probability that arises from the dot product will be employed to ascertain which pixels warrant attention. In other terms, the pixels that are observed are similar to these pixels.

Attention block is initially introduced in YOLOv10 and YOLOv11. In YOLO11, the attention process is represented as C2PSA. This block has two convolutional blocks positioned at the initial and the terminal. The outcome of the initial convolutional block is divided. The first segment of the results will be processed by multiple PSABlocks to ascertain which components necessitate attention. The outcomes from PSABlock will be combined, utilizing a concat block, with the second segment of the results.

PSABlock contains an Attention block and FFN (Feed Forward Network). FFN comprises two consecutive convolutional blocks. Nevertheless, the second convolutional block does not incorporate an activation function.

YOLOv10 similarly possesses modules that serve to identify regions of the image that necessitate attention, referred to as PSA (Partial Self Attention). This block is analogous to C2PSA; however, it features a more straightforward structure. It initiates with a convolutional block that bisects the results, with one half proceeding directly to the concatenation block whereas the other half is processed by the attention block and the convolutional block. At the block's end, it employs a convolutional block to process the concatenated results.

The PSA in YOLOv10 comprises a singular module featuring one attention layer and one feed-forward layer. C2PSA is an advanced module comprising multiple PSA blocks. C2PSA is more complex than PSA as it iteratively performs the PSA operation, thereby capturing more profound feature relationships. C2PSA efficiently captures deeper features without additional time expenditure, as the initial split output convolution splits the input into two pathways: shortcut and deep processing.

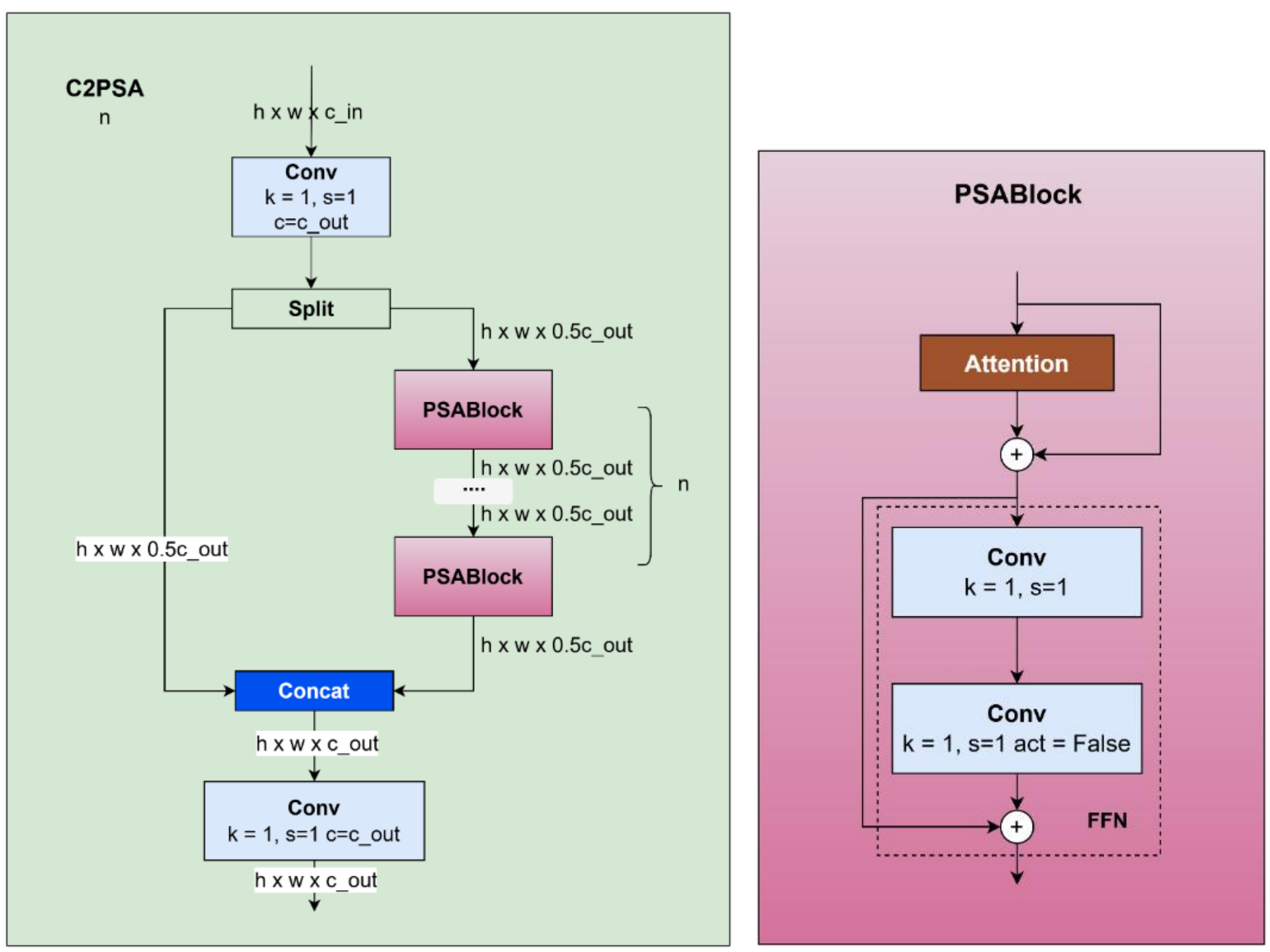

Figure 18: C2PSA and Attention Block

### 4.2.10 Detect Block

In YOLOv8 and YOLOv9, the detect block comprises two components, specifically the classification head and the regression head. Class probabilities are generated by the classification head, whereas bounding box coordinate predictions are generated by the regression head. Each head is constituted of two convolutional blocks, succeeded by a two-dimensional convolution layer.

In the YOLOv10 detect block, the classification head and regression head exhibit distinct architectures. This differentiation was necessitated by the disparity in computational costs. The classification head may be twofold larger than the regression head.

Nonetheless, the regression head plays a more pivotal role in ascertaining performance. Consequently, depthwise convolutions may be employed to optimize the classification head architecture, rendering it more lightweight. With this strategy, the number of model parameters is reduced. Figure 19 illustrates YOLOv8, YOLOv9 and YOLOv10 Detect block.

YOLOv10 introduces the concept of Consistent Dual Assignment for NMS-free Training. This concept integrates one-to-many and one-to-one approaches in model training, utilizing consistent matching metrics. This strategy employs two heads during training: one implements one-to-many assignment for robust supervision, whereas the other utilizes one-to-one matching. During inference, only the head utilizing one-to-one matching is employed, eliminating the necessity for NMS and improving inference efficiency without sacrificing performance. [4].

The detect block in YOLO11 uses depthwise convolutions in the classification head, which is nearly identical to YOLOv10. Nevertheless, YOLO11 incorporates a novel block, specifically DWConv. Figure 20 illustrates YOLO11 Detect block.

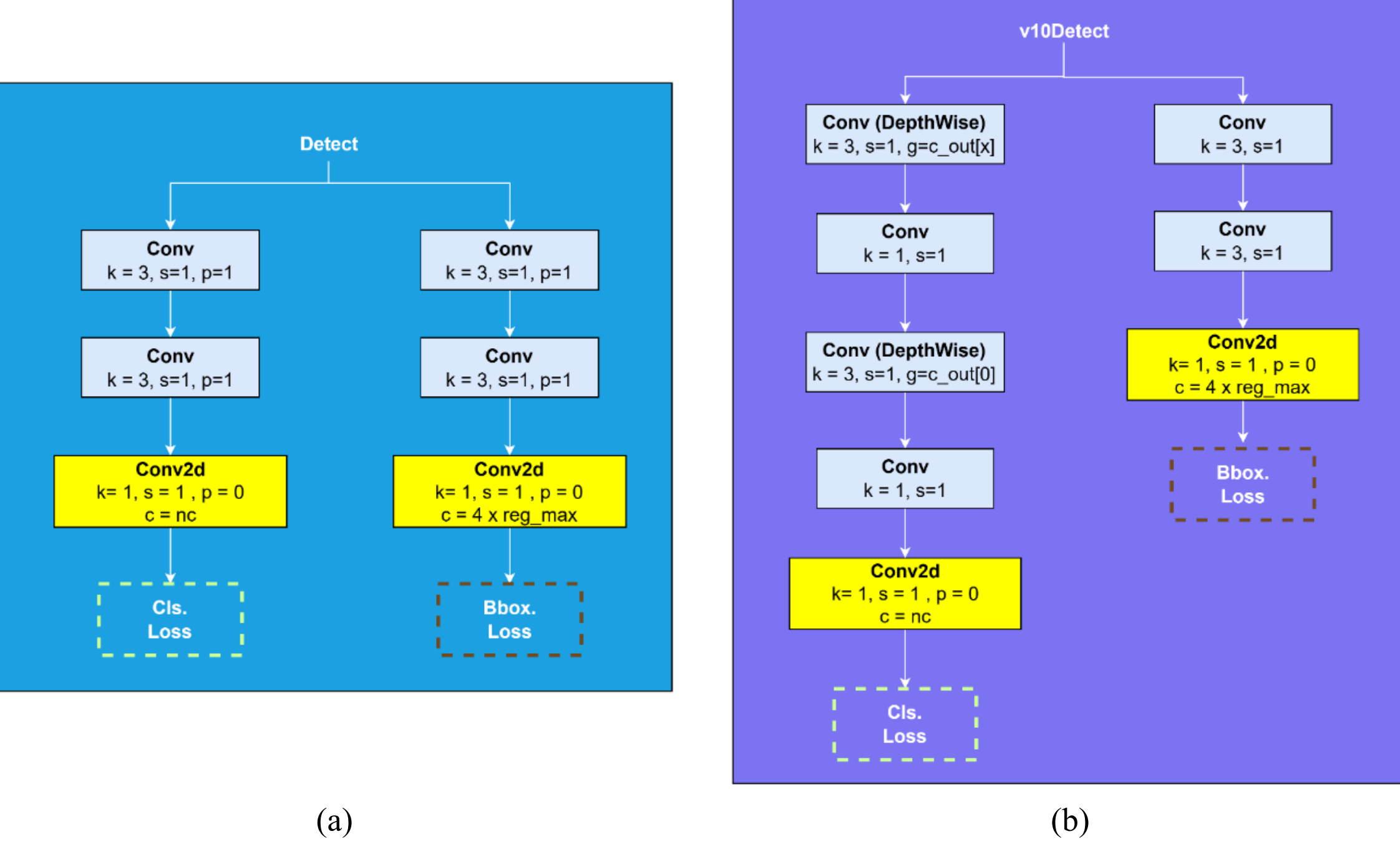

(a) (b)

Figure 19: (a) YOLOv8 and YOLOv9 Detect Block (b) YOLOv10 Detect Block.

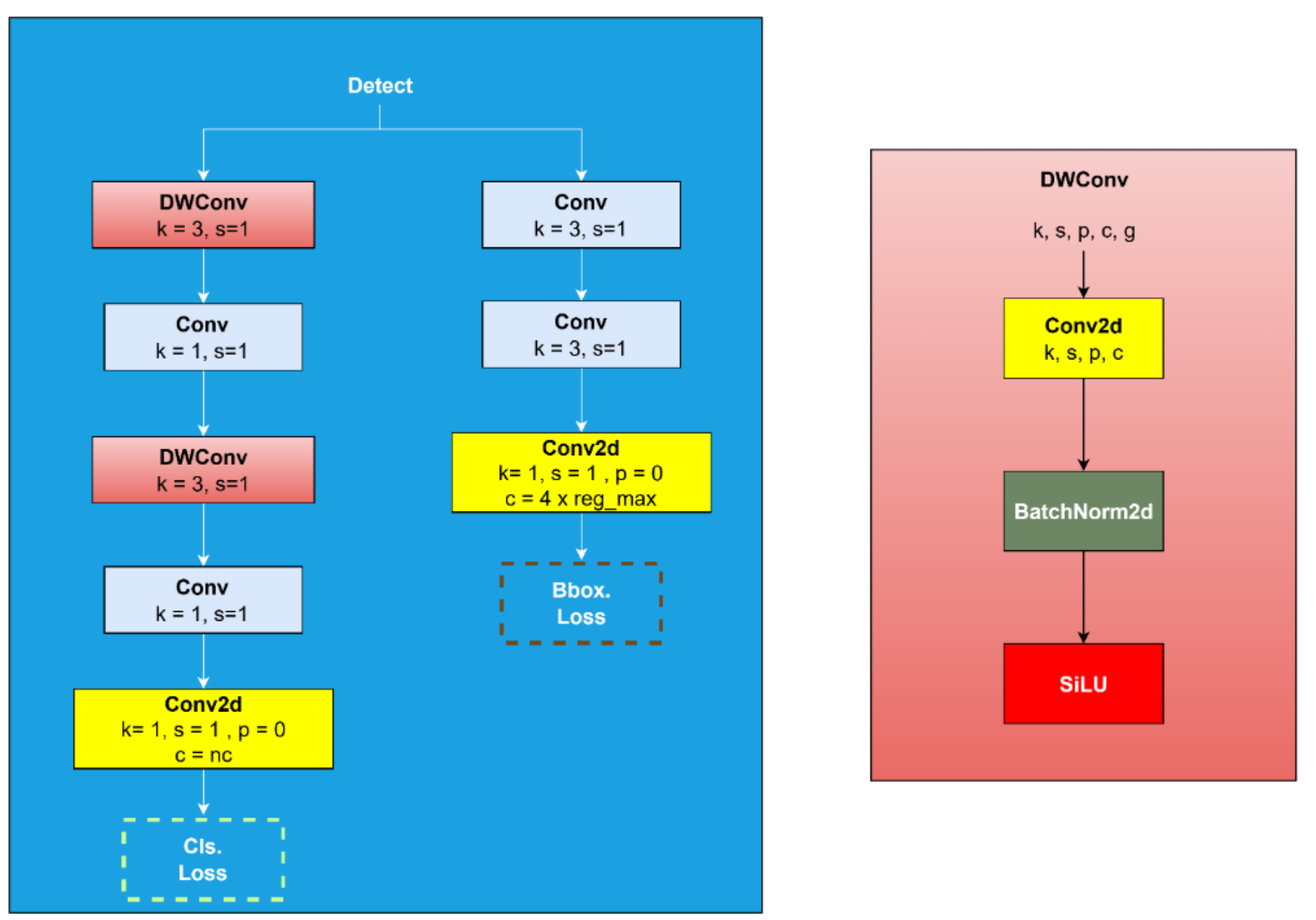

Figure 20: YOLO11 Detect Block

From the above comparison, we can draw a comprehensive comparison between the four YOLO versions as presented in Table 2.

Table 2: YOLO Comprehensive Comparison

| Aspect | YOLOv8 | YOLOv9 | YOLOv10 | YOLO11 |
|---|---|---|---|---|
| NMS | NMS | NMS | NMS-free | NMS |
| Anchor | Anchor-free | Anchor-free | Anchor-free | Anchor-free |
| Novelty | C2f | PGI | SCDown | C3k2 |
| | SPPF | GELAN | PSA | C2PSA |
| | | ADown | Conv(DepthWise) | DWConv |

### 4.3 An Illustration of How the Architecture Processes an Image

This illustration utilizes YOLO11l for demonstration purposes. We selected this model because YOLO11 is the latest version of YOLO, lacking an official architectural diagram and published paper. The l variant was selected due to its simpler calculations compared to the others.

The model is employed with a depth_multiple of 1, a width_multiple of 1, and a max_channels of 512 as presented in Table 3. The input image measures 640 x 640 pixels and comprises 3 channels. The initial input image is processed through Block 0, which is a convolutional block featuring a kernel size of 3 and a stride of 2. Applying a stride of 2 reduces the spatial resolution of the output to 320 x 320. The output channels are determined by multiplying the lesser value between the base output channels and max_channels by the width_multiple. In Block no. 0, the base output channels total 64, resulting in output dimensions of 320 x 320 x 64. The process is illustrated in Figure 21.

Table 3: Model Parameters

| Model Variant | depth_multiple | width_multiple | max_channels |
|---|---|---|---|
| YOLO11l | 1 | 1 | 512 |

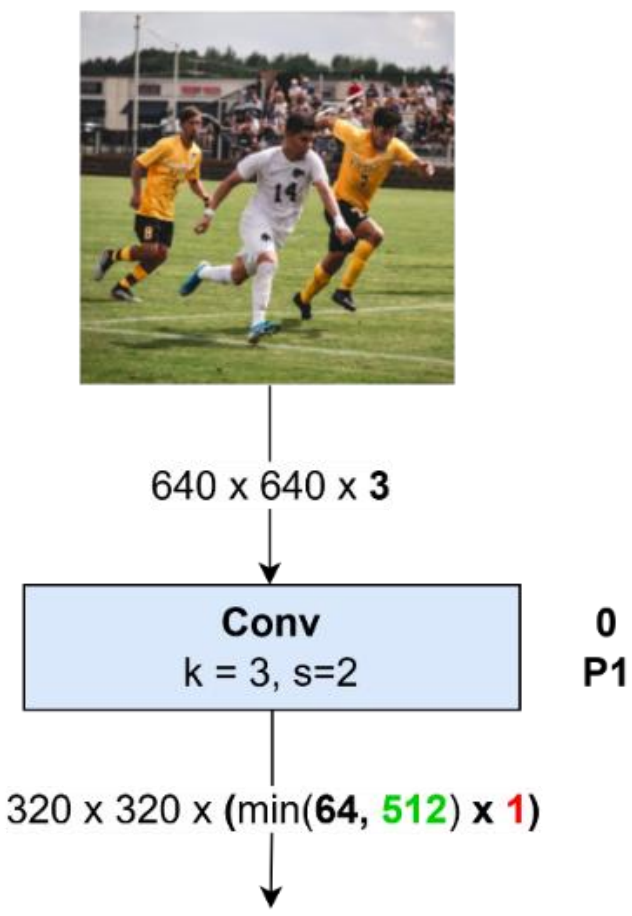

Figure 21: Input and Block Conv No. 0

The image is subsequently processed through Block 1, which is an additional convolutional block featuring a kernel size of 3 and a stride of 2. The spatial resolution is further reduced to 160 x 160, and the number of channels is adjusted to 128. The output from Block No. 1 is transmitted to Block C3k2 (Block No. 2). In this block, deeper feature extraction is conducted. The output maintains a spatial resolution of 160 x 160, whereas the number of channels increases to 256. The process is illustrated in Figure 22.

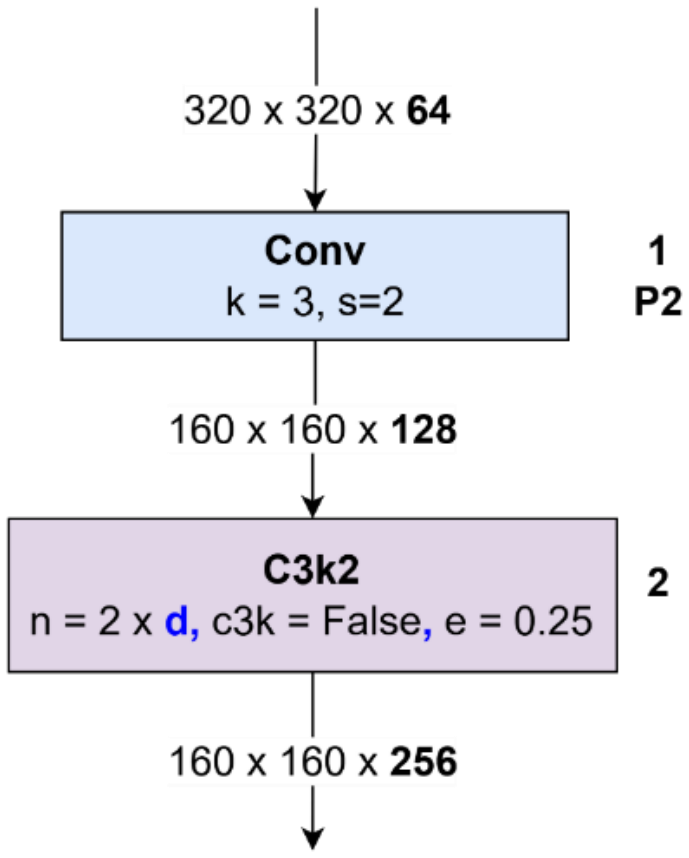

Figure 22: Block Conv No.1 and C3k2 No. 2

This process continues gradually through subsequent blocks in the Backbone, where the spatial size is progressively reduced to 80 x 80 (Block no. 3), 40 x 40 (Block no. 5), and 20 x 20 (Block no. 7), according to the configuration of convolutional layers and stride. Additionally, there are several C3k2 blocks (Block no. 4, 6, and 8) that are connected to the neck section. The process is illustrated in Figure 23.

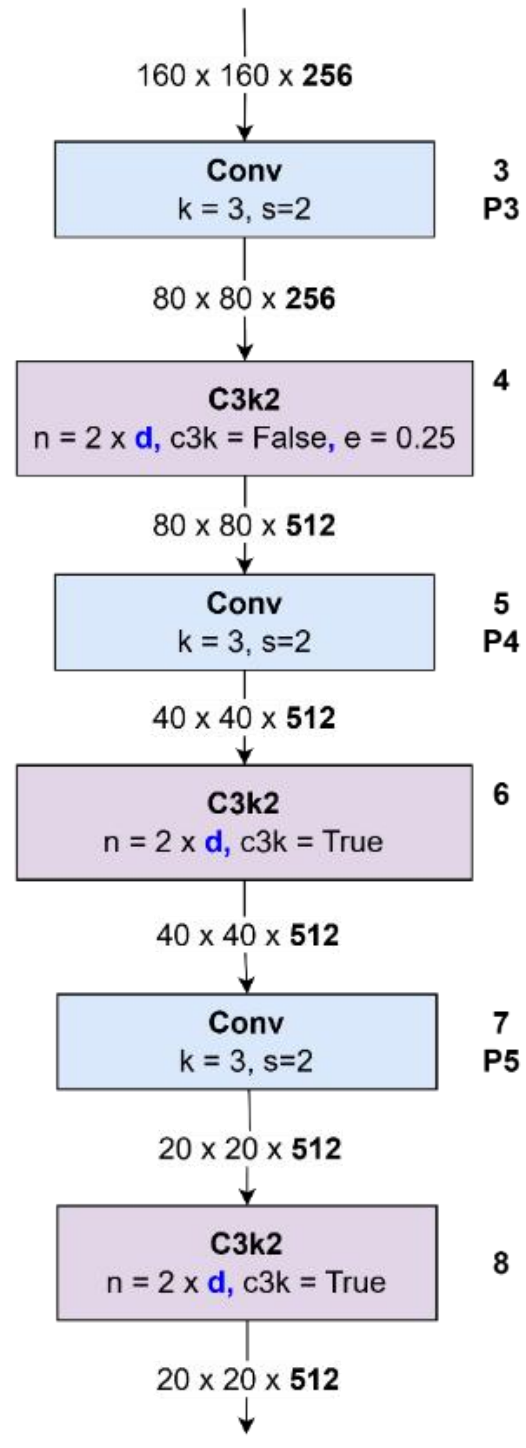

Figure 23: Block No. 3 - 8

The output from the backbone, which is the output of Block C3k2 no. 8, is 20 x 20 x 512. This output is transmitted to the neck, specifically to Block SPPF no. 9. The output remains 20 x 20 x 512. The SPPF enhances the efficiency of integrating spatial information. The process is illustrated in Figure 24.

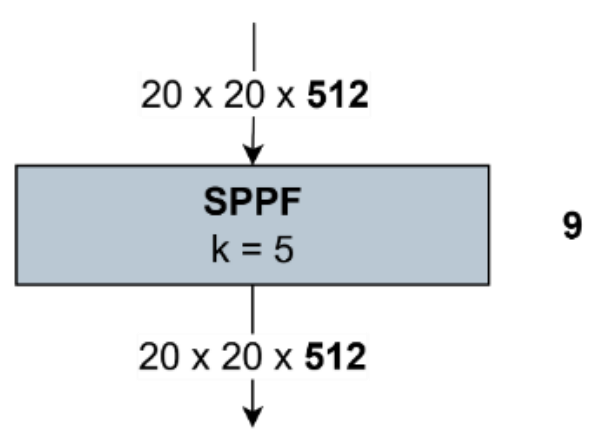

Figure 24: SPPF Block

Subsequently, it proceeds to Block C2PSA no. 9. This block is positioned exclusively at the lowest resolution stage. This is implemented to mitigate excessive computation arising from the complexity of self-attention. The output from this block is consistently 20 x 20 x 512. The process is illustrated in Figure 25.

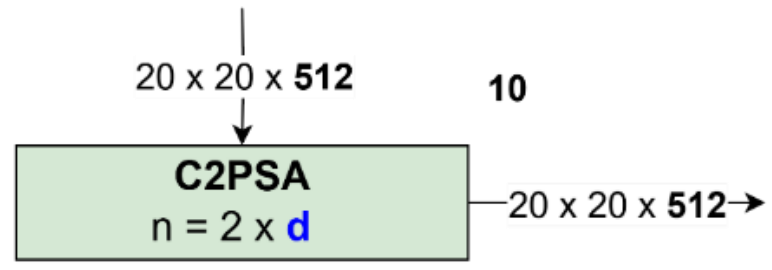

Figure 25: C2PSA Block

The output from the C2PSA block will be upsampled to 40 x 40 x 512, and this feature will subsequently be concatenated with the output from Block C3k2 no. 6 (40 x 40 x 512). In concatenation, the total number of channels is aggregated, whereas the resolution remains constant. The output from Concat (Block no. 12) is 40 x 40 x 1024. This procedure of upsampling and concatenation executes the Feature Pyramid Network (FPN), which seeks to integrate information from features of varying resolutions and levels of abstraction. The process is illustrated in Figure 26.

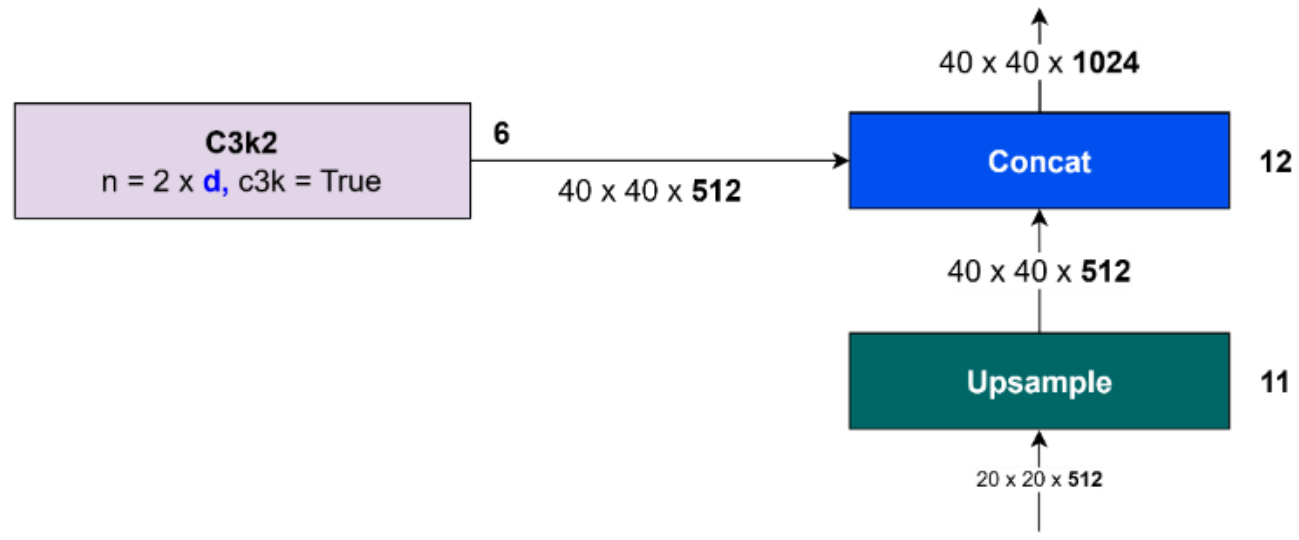

Figure 26: Upsampling and Concat

The process proceeds to Block C3k2 no. 13, yielding an output of 40 x 40 x 512. The output is subsequently upsampled to 80 x 80 x 512, and the feature is concatenated with the output from Block C3k2 no. 4 (80 x 80 x 512). The output from Concat number 15 is 80 x 80 x 1024. The process is illustrated in Figure 27.

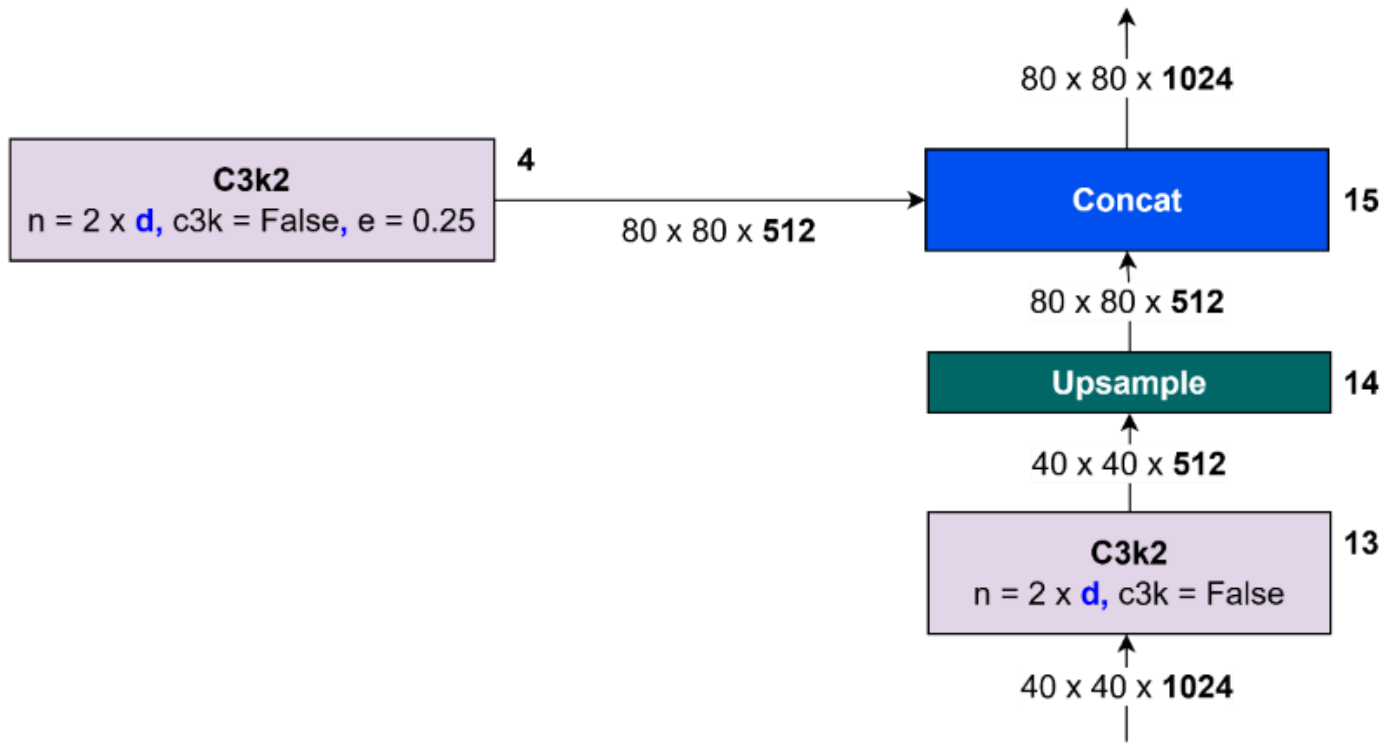

Figure 27: Concat Block No. 15

The output from Concat no. 15 is transmitted to Block C3k2 no. 16, yielding a feature map of 80 x 80 x 512. The output from this block serves as input for the Detect block, which is designed to identify relatively small objects within the image. The 80 x 80 dimensions are employed for identifying small objects, as the higher spatial resolution retains more details from the original image. The process is illustrated in Figure 28.

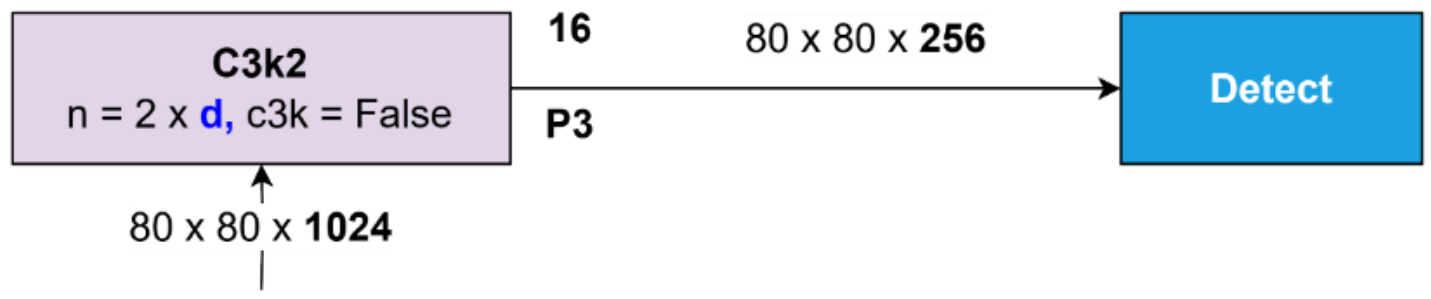

Figure 28: Detect Block for Small Object

The output from Block C3k2 no. 16 is subsequently transmitted to Block Conv no. 17. The spatial resolution decreases to 40 x 40, whereas the channel count remains at 256. Larger objects typically occupy a broader area within the image. Lowering the resolution allows for a more compact representation of the object's features, enhancing the model's capacity to capture patterns at the correct scale. The output from Block Conv no. 17 is concatenated with the output from Block C3k2 no. 14, yielding a feature map of dimensions 40 x 40 x 768. The process is illustrated in Figure 29.

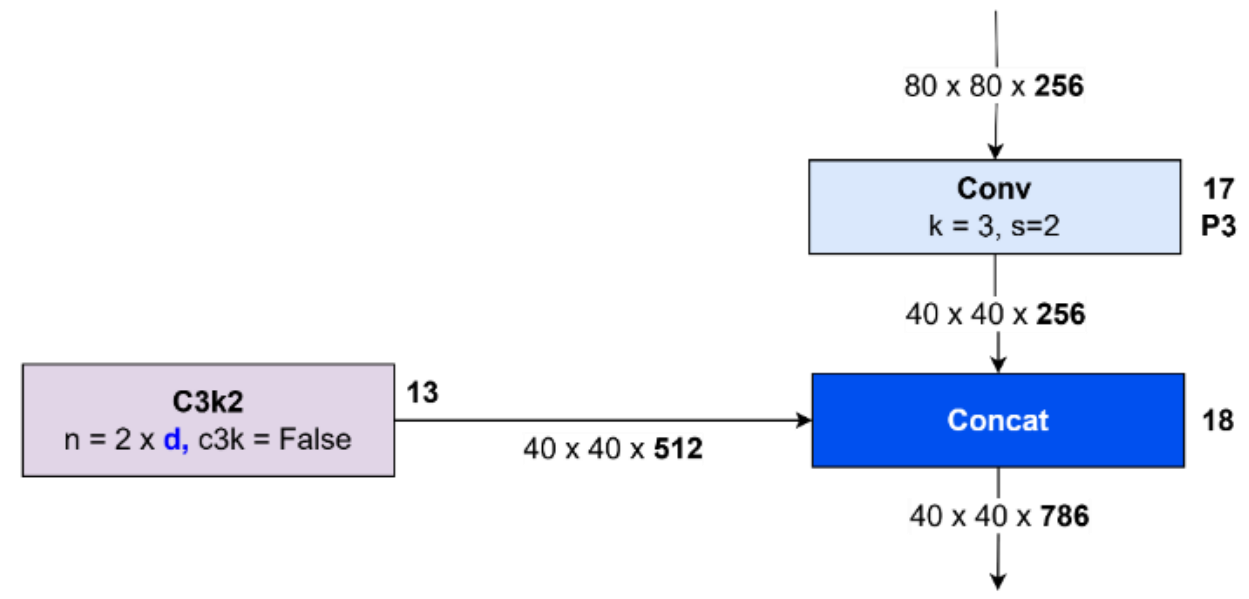

Figure 29: Concat Block No. 18

The output from Concat no. 18 is subsequently transmitted to Block C3k2 no. 19, resulting in a feature map of 40 x 40 x 512. The output from this block serves as input for the Detect block, which is specifically designed to identify medium-sized objects. The process is illustrated in Figure 30.

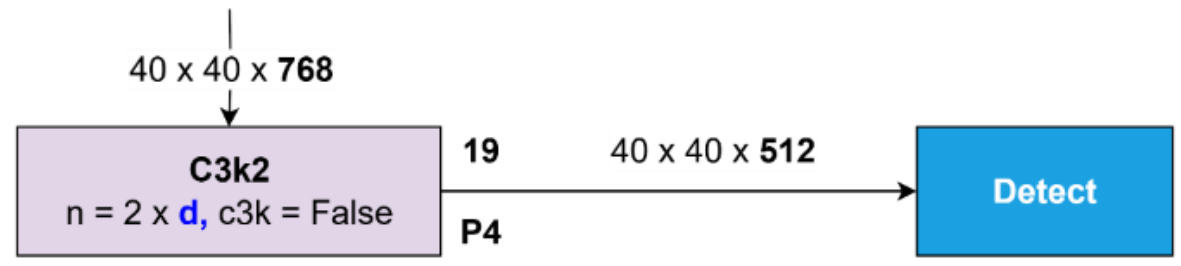

Figure 30: Detect Block for Medium Object

The output from Block C3k2 no. 19 is subsequently processed through Block Conv no. 20, resulting in a spatial resolution of 20 x 20, while maintaining 512 channels. The output from Block Conv no. 20 is concatenated with the output from Block C2PSA no. 10, yielding a feature map of dimensions 20 x 20 x 1024. The output from Concat no. 21 is subsequently transmitted to Block C3k2 no. 22, resulting in a feature map of dimensions 20 x 20 x 512. The output from this block serves as input for the Detect block, which is specifically designed to identify large objects. The process is illustrated in Figure 31.

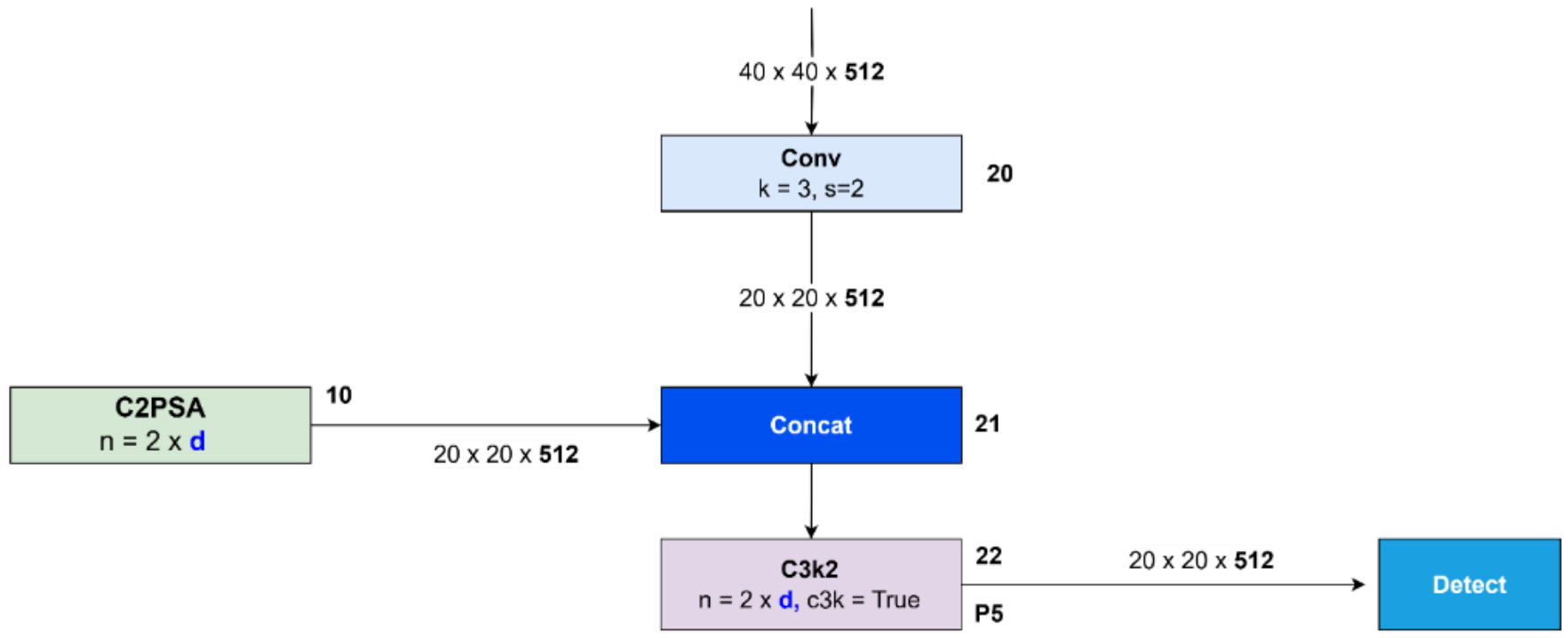

Figure 31: Detect block for Big Object

## 5 Conclusion

Our comprehensive in-depth analysis revealed that each version exhibits enhancements and refinements in architecture and feature extraction. However, we identified certain architectural blocks that are identical to the previous version. The absence of scholarly publications and official architecture diagrams presents challenges for researchers and practitioners in understanding the model's functionality. This issue also presents a challenge for future model enhancement. We also discovered that there is an evolutionary trend in YOLO, incorporating an attention layer beginning with YOLOv10. We urge future YOLO developers to provide an architectural diagram and a scholarly publication to accelerate researchers, practitioners, and developers in comprehending the model and enhancing its performance in the future.